%% file: axi_main.tex
\documentclass{article}
\usepackage{iclr2027_conference,times}
\iclrfinalcopy
\usepackage[utf8]{inputenc}
\usepackage[T1]{fontenc}
\usepackage{hyperref}
\usepackage{url}
\usepackage{booktabs}
\usepackage{amsfonts}
\usepackage{nicefrac}
\usepackage{microtype}
\usepackage{xcolor}
\usepackage{amsmath,amssymb,mathtools,amsthm}
\usepackage{graphicx}
\usepackage{float}
\usepackage{pgfplots}
\pgfplotsset{compat=1.17}
\usepgfplotslibrary{polar}
\usetikzlibrary{arrows.meta, positioning, calc, fit, backgrounds, matrix}

\newtheorem{proposition}{Proposition}

\input{axi_macros}
\input{axi_numbers}

\title{Context-Tower Conversion Preserves Generation While Freezing
Retains Knowledge: Low-Budget AR-to-Diffusion Conversion of MoE LLMs}

\author{%
  Wentao Lu\thanks{Corresponding author.}, Tianyu Zhu, Jesse Clark \\
  Celeris, San Francisco, CA \\
  \texttt{\{toby,alan,jesse\}@celeris.ai}
}

\begin{document}
\maketitle
\lhead{Preprint}

\begin{abstract}
Converting a pretrained autoregressive (AR) model to a diffusion language
model (dLLM) enables parallel generation without pretraining a new model.
Published conversion methods differ by roughly three orders of magnitude in
training data and have not been compared under a common protocol. We compare
two conversions of the same 30B Mixture-of-Experts (MoE) parent, holding the
corpus, supervised-token budget, trainable parameter set and evaluation
harness fixed, each under its own training recipe. The in-place model
updates a subset of the parent's weights using denoising and
representation-alignment losses; the frozen-tower model instead conditions
through cross-attention on a frozen causal copy of the parent. With 1\,B
training tokens, the frozen-tower model scores \qttHEten{} on \heval{}
\passat{10} against \dfiveHEten{} for the in-place model, an $11.6\times$
improvement. At the same budget it also keeps 95\% of the parent's GSM8K
score and 99\% of its \mmlupro{} score. A dense-parent experiment reproduces
the \heval{} separation. Within the two-tower design at about 500\,M tokens,
freezing the context tower retains substantially more \mmlupro{} performance
than training it, while both give similar observed \heval{} scores. Our
theoretical analysis establishes that both conversion classes contain an
exact sampler for the AR parent under a hard attention mask and
left-to-right commitment of one position per round. Under a shared loss,
freezing removes the gradient contribution through the context states.
Furthermore, evaluation protocol substantially affects a published
500\,B-token conversion's scores in both directions across tasks, while its
AR parent's scores vary by less than three points, so comparing dLLMs needs a
common protocol. These results show that, in the tested low-budget regime,
the frozen-tower configuration retains substantially more of the parent's
generation performance than in-place conversion. The construction applies to
any causal transformer with softmax attention.
\end{abstract}

\section{Introduction}\label{sec:intro}

A common route to a diffusion language model (dLLM) is to initialise from a
pretrained autoregressive (AR) model and continue training with a denoising
objective \citep{gong2025diffullama,ye2025dream}. Existing conversion studies
are difficult to compare. Their data budgets range from $\sim$1\,B tokens
\citep{wu2025fastdllm2} to 500\,B tokens \citep{rnd1report2025}, their recipes
vary across several coupled design choices, and their models have not been
evaluated under a common protocol.

We study low-cost AR-to-diffusion conversion under matched training and
evaluation. Within the tested regime, the frozen-tower configuration, which we call
context-tower conversion, retains most of the parent's generation
performance at budgets where in-place conversion retains little. This gain
comes from the whole configuration. A trainable-tower control gives
similar observed \heval{} scores with frozen and trainable towers, while
freezing retains substantially more \mmlupro{} performance
(\S\ref{sec:theory}). We fixed the evaluation gates, numerical
thresholds, and stopping criteria before each run and evaluated every model
with the same fixed harness. Our contributions are:

\begin{itemize}
\item An \emph{output-length-dependent failure mode} for in-place conversion
  (\S\ref{sec:inplace}). Across the masked-diffusion recipes, the converted
  models retain most of the parent's performance on knowledge benchmarks,
  while generation performance declines as output length increases.
\item A \emph{matched comparison of two conversion configurations}
  (\S\ref{sec:pair}). Two conversions of the same 30B MoE parent share
  corpus, budget, trainable parameter set and evaluation harness; one trains a
  subset of the parent weights under denoising and alignment losses, the other
  reads a frozen causal context tower \citep{reda2026twotower}. The frozen-tower model reaches
  \qttHEten{} on \heval{} \passat{10} against \dfiveHEten{}, a factor of
    $\gtrsim11$ (\S\ref{sec:dispersion} gives the resampled range).
\item An \emph{evaluation practice for cross-protocol comparisons}
  (\S\ref{sec:protocols}). We found that decoding protocol changes a
  converted model's score by up to $4.8\times$ while the AR parent's scores
  move by less than three points. Single-draw differences below three
  points may disappear or reverse when decoding is repeated
  (\S\ref{sec:dispersion}). We therefore compare on a fixed task set with
  the parent as a control, report sampled and deterministic decoding
  separately, and re-decode with three seeds.
\item A \emph{budget-efficiency comparison}. With 1\,B training tokens, the
  frozen-tower model reaches \rndoneRetLo--\rndoneRetHi\% of the \heval{}
  \passat{10} score of a
  published 500\,B-token conversion under that model's inference code.
\item A \emph{trainable-tower control} and a gradient-isolation analysis
  (\S\ref{sec:theory}). Two models trained with the same recipe to
  approximately 500\,M tokens, one with a frozen tower and one with a
  trainable tower, achieve similar observed \heval{} scores, but the
  frozen-tower model scores \ttMMLUPdropAbs{} points higher on \mmlupro{}. Under
  the same loss,
  freezing removes one gradient term. Both configurations can reproduce the
  parent under sequential sampling, and Appendix~\ref{app:witness} runs
  that sampler.
\item \emph{Two ablations of the tower--denoiser interface} (\S\ref{sec:interface}).
  Attach density is varied over 25--100\% of tower layers, and denoiser
  size over three donors at a fixed tower.
\end{itemize}

\paragraph{Scope.} We focus on MoE parents because freezing the experts leaves
only $\sim$3\% of the parameters trainable. The construction applies to any
causal transformer with softmax attention.

\section{Output-length-dependent degradation under in-place conversion}\label{sec:inplace}

We use \olmoe{} \citep{muennighoff2024olmoe}, an MoE with 1\,B active parameters and a
public pretraining mixture. Continuing on this mixture reduces distribution
shift during conversion. We test four in-place recipes: (i) full fine-tuning
with masked diffusion \citep{sahoo2024mdlm}; (ii) frozen experts with
layer-wise representation alignment to the frozen AR teacher
\citep{peng2026repralign}; (iii) block diffusion
\citep{arriola2025bd3lm,wu2025fastdllm2} with an exact prefix KV cache; and
(iv) uniform-state corruption with self-conditioning. Each arm uses
0.75--1\,B supervised tokens.

All masked-diffusion arms show the same dependence on output length
(Table~\ref{tab:inplace}; Appendix~\ref{app:length} measures the length axis). They retain $\approx\inplaceKnowledgeRet\%$ of the
parent's \mmlupro{} score. BoolQ remains near the parent, short extractive
answers degrade, and JSON validity falls from 99.6\% to 0.0--1.8\%. Code
generation also degrades sharply: MBPP is 0.0 for every masked-diffusion arm,
and \heval{} remains far below the parent.

The uniform-state arm, evaluated with its native entropy-bound sampler, also
loses single-token performance (BoolQ 12.8). That score is far below the
62\% majority-class baseline. The harness counts an answer without a
yes/no word as wrong, and the arm's JSON outputs show token-level
corruption, so a failure to emit a valid answer token is a possible
explanation. We did not count invalid BoolQ answers. Applying the same sampler to the
block-model weights reduces \mmlupro{} from 15.4 to 0.6, although those
weights were trained with masked diffusion. The additional loss therefore comes from the decoder, because the two
columns share the same trained weights and the same training objective. The
small-parent BoolQ scores are close to the $\approx62\%$ majority-class
baseline and therefore provide weak evidence for single-token retention. The
30B experiment in Figure~\ref{fig:radar} provides the stronger control.

The output-length profile is unchanged by a $25\times$ change in trainable
parameters, a different attention layout, or additional training within the
low-budget regime. Repeating 0.8\,B tokens and training on 1\,B fresh tokens
produce flat trajectories at matched steps. A published 500\,B-token conversion of the 30B parent
\citep{rnd1report2025} reaches parity on \heval{} under its own inference
code (\S\ref{sec:protocols}), consistent with more training reducing the
deficit. It does not isolate budget.

\begin{table}[ht]
\caption{\textbf{In-place conversion of \olmoe{}.} Each arm uses
0.75--1\,B supervised tokens and the fixed evaluation harness. The lower rows
report the structured-output suite (BoolQ, SQuAD-short EM and CoNLL-JSON
validity). We added this suite after the phase gates and ran it on the AR
base and the three arms then live. The full fine-tuning arm had been
superseded by (ii) at the first gate, and the entropy-decoder column reuses
(iii)'s weights and was scored on the primary suite only. $^{\S}$The same block-model
weights evaluated with the entropy-bound decoder; \mmlupro{} decreases from
15.4 to 0.6. Each cell uses one sampling run; \passat{10} uses $n=20$
samples. Sampling seeds were not fixed.}
\label{tab:inplace}
\centering\small
\begin{tabular}{lcccccc}
\toprule
 & AR base & (i) full-FT & (ii) +align & (iii) block &
 (iii)$\times$ent.$^{\S}$ & (iv) uniform \\
\midrule
\heval{} \passat{10} & \olmoeHEten & \ponHEten & \ptwoHEten &
  \pthreeHEten & 0.00 & 0.00 \\
GSM8K & 12.59 & 1.67 & 2.12 & 1.82 & 0.15 & 1.59 \\
MBPP & 18.60 & 0.00 & 0.00 & 0.00 & 0.00 & 0.00 \\
\mmlupro{} & 19.80 & 15.20 & 15.20 & 15.40 & 0.60 & 8.40 \\
\midrule
BoolQ (\%) & 63.2 & --- & 63.4 & 62.0 & --- & 12.8 \\
SQuAD-short EM (\%) & 67.4 & --- & 44.0 & 22.0 & --- & 0.6 \\
JSON validity (\%) & 99.6 & --- & 1.8 & 0.0 & --- & 9.6 \\
\bottomrule
\end{tabular}
\end{table}

\section{The matched pair}\label{sec:pair}

\paragraph{Design.} \dfive{} and \qtt{} use the same parent (\parent{}
\citep{yang2025qwen3}; \heval{} \passat{10} \parentHEten), corpus,
supervised-token budget (\qttBudget{}), bit-identical data manifest,
trainable parameter set (0.919\,B parameters), optimiser and schedule, and
evaluation harness. They differ in context architecture, alignment loss,
block-size schedule and corruption procedure (Appendix~\ref{app:mech}): the
comparison evaluates two conversion configurations and does not isolate
freezing alone. \qtt{} reads a frozen,
layer-aligned copy of the parent through cross-attention
(Figure~\ref{fig:arch}); \dfive{} computes the same context with the weights
being retrained. The tower keys, rotated at their absolute positions, and the tower values
are concatenated into the denoiser's existing attention, so the attachment
adds no trainable parameters.

\input{figures/axi_fig_arch_twotower}

\paragraph{Results (Figure~\ref{fig:radar}).}
The in-place arm reaches \heval{} \passat{10} \textbf{\dfiveHEten}, within the prespecified insufficiency band ($\leq10$). Its score remains flat from
250\,M to 1\,B tokens: $4.96 \to 4.26 \to 4.38 \to 6.19$. It also scores
\dfiveMBPP{} on MBPP and \dfiveGSM{} on GSM8K. The 30B parent therefore shows
the same low-budget in-place deficit as \olmoe{}.

The frozen-tower arm reaches \textbf{\qttHEten} on \heval{} \passat{10}, a
$\gtrsim11\times$ improvement and \ebOneRetLo--\ebOneRetHi\% of the parent
score (seed floor to committed draw). The committed scores were obtained on each arm's training host. When both arms were evaluated on the same Blackwell host using three decoding seeds, their mean scores were \ebTwoQttMean{} for the frozen-tower model and \ebTwoDFiveMean{} for the in-place model (Appendix~\ref{app:disp}).
It scores
\qttGSM{} on GSM8K, \qttMMLUP{} on \mmlupro{} and \qttMBPP{} on MBPP, or
95\%, 99\% and 72\% of the parent. Its JSON validity is
\qttJSONvalid\%, compared with \dfiveJSONvalid\% for the in-place arm and
0.0--9.6\% for the small-parent in-place arms.

The in-place arm still retains 82\% of the parent \mmlupro{} score
(\dfiveMMLUP{} against \parentMMLUP{}) and does better on short and structured output than the
small-parent in-place arms. Its SQuAD EM exceeds the parent score (\dfiveSquad{} vs.\
\parentSquad), but this difference comes from output formatting. The parent over-generates
under the short-answer prompt,
whereas the converted model emits length-bounded blocks. The same effect
accounts for the lower SQuAD score of the 30B parent relative to the 4B
parent.

Long-form generation nevertheless remains poor for the in-place arm. Because the two arms optimise the same parameter set, a difference in
trainable parameter count does not explain the gap. The small-parent full-fine-tuning arm, with $25\times$
the trainable count, shows the same profile (\S\ref{sec:inplace}). On held-out data each arm reduces its own diffusion objective by about
twelve nats from initialisation (\elboInplace{} and \elboFrozen{} at 1\,B
tokens; Appendix~\ref{app:mech}). The objectives differ so these losses do
not measure relative convergence.

\paragraph{Relation to prior two-tower work.}
\citet{reda2026twotower} introduced the frozen-tower architecture and report
\twotowerRetention\% quality retention at \twotowerSpeedup$\times$
throughput after training on \twotowerBudget{} tokens. Their Mamba-hybrid MoE
parent does not permit a directly matched in-place conversion because most
layers are recurrent and expose no causal attention mask to replace.

\paragraph{Replication across seeds and hardware.} We trained a second
\qtt{} seed with a new data order and corruption sequence on different
hardware. Two bit-parity tests found no measurable difference between the
hardware implementations. At 500\,M tokens, the seeds differ by less than
half a percentage point on \heval{} \passat{10} (\qttSTwoHEtenB{} vs.\
\qttHEtenB), within the prespecified $\pm5$-point band. Their \mmlupro{}
scores are identical (\qttSTwoMMLUP{}), and the second seed reaches
\qttSTwoJSONvalid\% JSON validity.

At 250\,M tokens, the \heval{} \passat{10} scores differ by
$\sim$\rTwoSpreadMid{} percentage points and fall outside the prespecified band. This checkpoint fails the mid-training replication criterion. By
500\,M tokens the difference disappears, while knowledge metrics remain
within one point. For this model and training setup,
mid-training \heval{} differences below $\sim10$ percentage points fall
within the observed \emph{training}-seed variation; MBPP and knowledge
benchmarks were more stable. Resampling the decode on both arms places the
500\,M comparison inside \emph{sampling} dispersion as well: three draws give
\ebTwoSeedOneMean{} for the first seed and \ebTwoSeedTwoMean{} for the
second, against a range of \ebTwoSeedTwoRange{} on the second arm alone. The
two seeds agree at 500\,M with no resolvable ordering. We did not train a second in-place seed. The in-place profile instead
appears in six arms across three parents: the four recipes in
\S\ref{sec:inplace}, \dfive{} on the 30B MoE, and \dpi{} on the dense parent.
The 65-point pair gap is much larger than the observed frozen-tower seed
variation, but does not replace replication of the in-place arm.

\paragraph{Sampling dispersion.}\label{sec:dispersion}
All scores other than the greedy rows come from sampling, and the committed
values are single draws. To see how far one draw can move, we decoded five
committed cells again with three decode seeds each, leaving the training and
the checkpoints untouched. Greedy decoding gave the same score in every cell.
Taken within one seed and one host, the pair ratio ranges from
\ebTwoRatioBand$\times$. Almost all of that width comes from the in-place
arm, whose three seeds span \ebTwoDFiveRange{} points around a mean of
\ebTwoDFiveMean{}. Two comparisons
showed single-draw differences of less than two points. When we re-decoded
both arms with three seeds, the sign of each difference flipped. So we mark
single-draw differences below three points as unresolved and draw no
conclusion from them. Appendix~\ref{app:disp} gives the
per-cell tables and the cross-host check behind the comparisons in
\S\ref{sec:protocols} and \S\ref{sec:throughput}.

\input{figures/axi_fig_radar}

\paragraph{Dense-parent replication.} We repeated the matched comparison with a dense parent (\dparent). We kept both configurations, the corpus, the attachment mechanism and the
evaluation harness fixed and changed the parent family
(Appendix~\ref{app:dense}, Table~\ref{tab:dense}). After \dpsBudget{} supervised tokens, the
frozen-tower model reaches \dpsHEten{} on \heval{} \passat{10}, or
\dpsRetPct\% of the parent score (\dparentHEten). This exceeds the prespecified retention floor of 45\%. It also remains close to the parent on
\mmlupro{} (\dpsMMLUP{} vs.\ \dparentMMLUP), BoolQ (\dpsBoolq{} vs.\
\dparentBoolq), SQuAD EM (\dpsSquad{}), and JSON validity
(\dpsJSONvalid\%).

The matched in-place arm uses the same corpus, budget, and seed. Its MLPs and
embeddings are frozen, while its attention layers and normalisation
parameters are trained; this is the same trainable set as the frozen-tower
denoiser. It reaches \dpiHEten{} on \heval{} \passat{10} and is already flat
at 250\,M tokens (\dpiHEtenA{}), within the prespecified ceiling of \dpiCeil{} (ratio \dpiRatio, or \dpiRetPct\% of the parent). It scores
\dpiMBPP{} on MBPP and \dpiJSONvalid\% on JSON validity while retaining
\dpiMMLUPRet\% of parent \mmlupro{}. The dense comparison reproduces the
length-dependent deficit at \densePairSep$\times$ separation on
\heval{} \passat{10}.

\paragraph{Comparison with a 500\,B-token conversion.} \rndone{} uses the same parent and a closely
matched freezing recipe, with a training budget of \rndoneBudget{}. Under its
native inference code and sampled protocol, it reaches parent parity on
\heval{} \passat{10} (\rndoneHEtenParity{} vs.\ \parentHEten), with
\rndoneHEtenParityCI{} across two disjoint seed sets. Relative to this
endpoint, \qtt{} reaches \rndoneRetLo--\rndoneRetHi\% of the score with
1\,B supervised tokens.

The comparison uses \rndone{}'s native decoder, which is favourable to the
baseline (\S\ref{sec:protocols}). Under greedy decoding, \qtt{} exceeds \rndone{} at every measured budget.
The sampled protocol, where the baseline is strongest, is compared in
\S\ref{sec:protocols}. Its
\heval{} greedy \passat{1} scores are \qttHEgreedyA{}, \qttHEgreedyB{}, and
\qttHEgreedy{} at 250\,M, 500\,M, and 1\,B tokens, all read on one Hopper
host lineage, compared with \rndoneHEgreedyParity{} for \rndone{} under its
native inference code. The greedy score falls by 2.4 points from 500\,M to 1\,B while
sampled \passat{10} rises from \qttHEtenB{} to \qttHEten{}. Greedy has no
decode dispersion (Appendix~\ref{app:disp}), so the fall is a property of the
checkpoint and is reported as measured.

\section{The role of frozen context}\label{sec:theory}

\paragraph{The trainable-tower control.} The frozen-tower recipe is rerun
with the tower's attention, router and norms trainable, at the same corpus,
data order and training seed (42) and approximately the same 500\,M-token
budget, with the bands fixed before the run (Appendix~\ref{app:prereg}).
Table~\ref{tab:control} gives the result. The
two arms have similar observed \heval{} scores, and the paired interval
includes both improvements and declines of about six points. \mmlupro{} is
\ttMMLUPdropAbs{} points lower with the tower trained, an interval that
excludes zero, and the gap is already present at 250\,M. Training the tower
also doubles the trainable count from 0.92\,B to 1.84\,B under the learning
rate and schedule set for the smaller set (Appendix~\ref{app:theory}), so
part of the \mmlupro{} drop may come from optimising the larger set with
those hyperparameters. This comparison does not separate that from movement
of the context representation. Together with \S\ref{sec:pair}, these
comparisons show a generation
advantage for the tested two-tower configuration, with no resolved \heval{}
advantage from freezing at about 500\,M tokens. Freezing gives higher
\mmlupro{} performance under the shared training schedule. Whether
separating context computation from denoising explains the generation
advantage is untested. One reading of Appendix~\ref{app:mech} is that
in-place training makes one set of weights do two jobs, causal context and
block denoising, and the context maps move away from the parent's. The
trainable tower's weights also change while its observed \heval{} score
stays similar, so weight change alone does not account for the in-place
loss.

\begin{table}[t]
\caption{\textbf{Trainable versus frozen context tower.} Same recipe,
corpus, data order and training seed; frozen = the committed seed-1
checkpoint at 4,960 steps (about 490\,M tokens), trainable at 5,059 steps
(500.08\,M). Differences are trainable minus frozen. $^{\dagger}$Committed-score
difference; the paired re-decode comparison gives $-0.33$ [$-6.13$,
$+5.26$] (Appendix~\ref{app:theory}). \mmlupro{} interval:
independent-binomial approximation (Appendix~\ref{app:theory}).}
\label{tab:control}
\centering\small
\begin{tabular}{lcc}
\toprule
 & \heval{} \passat{10} & \mmlupro{} \\
\midrule
frozen tower & \qttHEtenB{} & \qttSTwoMMLUP{} \\
trainable tower & \ttHEten{} & \ttMMLUP{} \\
difference & $\ttHEdiff{}$$^{\dagger}$ & $\ttMMLUPdiff{}$ [$\ttMMLUPdiffLo{}$, $\ttMMLUPdiffHi{}$] \\
\bottomrule
\end{tabular}
\end{table}

\paragraph{What the propositions add.} With
parent parameters $\theta^*$, per-layer key--value maps
$\mathrm{KV}_{\theta^*}(c)$ for a committed prefix $c$ and AR conditional
$\pi^*$, the frozen-tower class is
$\mathcal{G}=\{q_\phi(\cdot\mid z,\mathrm{KV}_{\theta^*}(c))\}$, with the
context states fixed; the in-place class has the same form with
$\mathrm{KV}_W(c)$ from the trained weights.

\begin{proposition}[Parent realisability with a frozen context]
\label{prop:witness}
Assume that attention uses a hard visibility mask, so a query at position $i$
places zero weight on masked positions and on visible positions after $i$.
Then
$\mathcal{G}$ contains a sampler $q_{\phi^\dagger}$ with parent-copied weights
and masked-key suppression. Under a left-to-right schedule that unmasks one
position per round, this sampler exactly reproduces the parent's AR
factorisation:
$p_{\phi^\dagger}(x_B\mid c)=\prod_i\pi^*(x_i\mid c,x_{B,<i})$.
\end{proposition}

The proof (Appendix~\ref{app:theory}) is an induction over layers. The
in-place class contains the same sampler at $W=\theta^*$, so realisability
does not separate the arms. We built this sampler, the witness, and ran it
(Appendix~\ref{app:witness}): it agrees closely with the parent's
greedy score at one position per round, and the trained checkpoint loses its score under the same
decoding rules, so this existence result does not establish that the
trained model preserves the parent under its native sampler. For the loss $L(\psi)=\mathbb{E}\,\ell(q_\psi(z,\mathrm{KV}_\psi(c)))$
the chain rule splits the in-place gradient into an emission term and a
context term. The frozen-tower gradient is the emission term alone, and at
$W=\theta^*$ the two emission terms coincide (Proposition~\ref{prop:decouple},
Appendix~\ref{app:theory}). Freezing removes the context term. After 1\,B
tokens the in-place arm's context maps sit at mean relative distance
\driftKmean{} and \driftVmean{} from the parent's (Appendix~\ref{app:mech}).

\section{Cross-protocol evaluation}\label{sec:protocols}

We evaluate two decoding protocols on a fixed task set and set the
published scores beside them (Table~\ref{tab:protocols}). The AR parent
moves by less than three points across the three columns. Between our decode
and the baseline's own code, the largest move is \parentParityMaxDelta{}
points, on \heval{} \passat{1} (\parentHEone{} against
\parentHEoneParity{}), which Table~\ref{tab:protocols} does not show.
This three-way agreement suggests that task formatting and harness
implementation are not the main source of the dLLM differences. The scores of \rndone{}, however, move by task-dependent amounts in both
directions across protocols. Under its native sampled protocol, it reaches parent parity
on \heval{} (\rndoneHEtenParity{}; sampled \passat{1}
\rndoneHEoneParity{}), whereas our original decoder gives a score lower by a
factor of $4.8$. Under its native greedy mode, the score falls to
\rndoneHEgreedyParity{} on greedy \passat{1}, a $4.6\times$ decrease at the
same metric. It also falls below our near-greedy decoder on every
deterministic task, by $2.2\times$ on GSM8K, so the disagreement runs in
both directions.

This behaviour is consistent with temperature changing the unmasking
order. Temperature scales the logits
before the entropy-ordered unmasking schedule, so it affects both sample diversity and unmasking order. We also ran \rndone{} under \emph{its own} code at temperature 0.01, which
separates an implementation difference from the temperature setting itself.
It scores \rndoneGSMTemp{} on GSM8K, \rndoneMBPPTemp{} on MBPP and
\rndoneMMLUPTemp{} on \mmlupro{}. These are within 1.6 points of our own
0.01 decode on all three tasks, and 5.6 to 11 points above its own greedy
mode. So most of the gap on the deterministic tasks closes once the
baseline's own code moves off greedy. The published sampled scores
are still 3.4 to 3.8 times higher than these, but that comparison also
differs in decoding protocol and evaluation harness. Cross-protocol
dLLM comparisons should therefore include parent-agreement diagnostics and
report sampled and deterministic decoding separately.

\begin{table}[t]
\caption{\textbf{Cross-protocol evaluation.} Two implementations on our task
set, beside the published scores from the authors' harness. Parent
scores agree across all three implementations, whereas \rndone{} varies
strongly with decoding mode. $^{*}$Parent results reproduced from
\citet{rnd1report2025}. $^{\ddagger}$Published MBPP and GSM8K results use
sampled decoding on the authors' harness; the parity cells use greedy
decoding on our task set. For \rndone{}, sampled \passat{1} changes from
\rndoneHEoneParity{} to \rndoneHEgreedyParity{} under native greedy
decoding. The $T{=}0.01$ rows are read in the text; on \heval{}, \passat{10}
collapses to \passat{1} there because $n{=}20$ near-deterministic draws buy
no diversity. ``n/r'' marks a cell the source does not report.}
\label{tab:protocols}
\centering\small
\resizebox{\linewidth}{!}{%
\begin{tabular}{lccc}
\toprule
 & published & our decode & parity (their code) \\
\midrule
\parent{} MBPP / GSM8K (diagnostic) & $74.2^{*}$ / $85.2^{*}$ & \parentMBPP{} / \parentGSM & \parentMBPPParity{} / \parentGSMParity \\
\parent{} \heval{} \passat{10} / greedy & n/r & \parentHEten{} / \parentHEgreedy & \parentHEtenParity{} / \parentHEgreedyParity \\
\rndone{} \heval{} \passat{10} (sampled) & $\sim$90s & \rndoneHEtenOurs & \textbf{\rndoneHEtenParity} \\
\rndone{} \heval{} greedy & n/r & 17.68 & \rndoneHEgreedyParity \\
\rndone{} MBPP / GSM8K & \rndoneMBPPPub{} / \rndoneGSMPub{} (sampled)$^{\ddagger}$ & \rndoneMBPPOurs{} / \rndoneGSMOurs{} (near-greedy) & \rndoneMBPPParity{} / \rndoneGSMParity{} (greedy) \\
\rndone{} MBPP / GSM8K / \mmlupro{} at $T{=}0.01$ & n/r & \rndoneMBPPOurs{} / \rndoneGSMOurs{} / \rndoneMMLUPOurs & \rndoneMBPPTemp{} / \rndoneGSMTemp{} / \rndoneMMLUPTemp \\
\bottomrule
\end{tabular}}
\end{table}

\section{Throughput and memory}\label{sec:throughput}

After 1\,B training tokens, \qtt{} commits \qttTokPerFwd{} tokens per forward.
It does not show the sharp-commit behaviour of \twotower{}
\citep{reda2026twotower}. A
\ssixteenBudget-token adaptation stage reduces the canvas block size $S$ from
32 to 16. The first-round commit fraction increases on every task, from
\qttFirstCommit\% to \ssixteenFirstCommit\% on \heval{}, while \passat{10}
reads \ssixteenHEten{}, above the prespecified floor of 65. The stage shows
no detected cost to quality: resampling both arms on three decode seeds puts
the adapted model at \ebTwoSsixteenMean{} against \ebOneQttMean{} for the
$S{=}32$ arm, both read on Hopper hosts, a difference of \ebTwoStageDelta{}
points in the adapted arm's favour with a paired interval of $[-1.9,+5.8]$
(Appendix~\ref{app:disp}). The $S{=}16$ arm is the $S{=}32$ model trained for a further
\ssixteenBudget{} tokens, so the comparison mixes the smaller block with the
extra budget and cannot separate them.

At batch size 1, the frozen-tower model runs at \latOneLo$\times$ the
throughput of the served AR baseline at $L=64$ and \latOneHi$\times$ at
$L=256$. Before the engineering pass described below, the same model ran at
\latOneBaseLo$\times$ and \latOneBaseHi$\times$ on the same hardware. Both
are measured on a pinned 20-prompt protocol, so the pass gained
\latOneGainHi$\times$ at $L=64$ and \latOneGainLo$\times$ at $L=256$. The
gain comes from an engineering pass on the canvas
forward, which took \latOneSinkShare\% of wall-clock time at $L=256$
beforehand. Static shapes and graph capture cut that step from
\latOneStepBefore\,ms to \latOneStepAfter\,ms in isolation, a factor of
\latOneStepGain. With capture cost and per-call synchronisation counted, the
same step gains \latOneStepGainProtHi$\times$ at $L=64$ and
\latOneStepGainProtLo$\times$ at $L=256$. All optimised paths are bit-identical to the
reference decoder. Faster variants that failed this test are excluded.

The tower and denoiser alias their frozen expert and embedding tensors. The resident weight footprint is therefore the parent model plus
\towerOverheadGB\,GB; activations and caches are additional.

The remaining throughput deficit is not explained by weight traffic. For a
256-token completion, the converted model streams \latOneTrafficQtt\,TB of
expert weights, compared with \latOneTrafficAR\,TB for served AR. It moves
\latOneTrafficPct\% less weight data but remains slower. The estimated
memory-bandwidth lower bound for the canvas forward is \latOneHBMFloor\,ms,
compared with the measured \latOneStepAfter\,ms. The profile therefore points
to small-batch MoE gather and launch overhead as the main optimisation
targets. We performed this optimisation after the architecture experiments,
so the quality results were obtained with a fixed inference stack. All
throughput numbers are at batch size 1. Larger batches may spread the MoE
gather overhead over more tokens. The results establish training-budget
efficiency and quality retention, but not an end-to-end serving advantage.

\section{Attach density and denoiser size}\label{sec:interface}

The denoiser connects to the tower through two settings: which tower layers
it reads, and how many parameters it has. We call these two settings the
interface. In every arm so far the denoiser read every tower layer and was a
full copy of the parent. This section changes one setting at a time and
holds the tower, recipe, corpus, budget and harness fixed.

\paragraph{Attach density.} Four arms read different numbers of tower
layers, each trained for 250\,M supervised tokens with the same data
order (Table~\ref{tab:interface}). The attach sets are nested and listed in
Appendix~\ref{app:interface}. The quarter and full arms were trained and scored on Hopper hosts and the
half and three-quarter arms on a Blackwell host (Appendix~\ref{app:repro}).
The quarter-to-full and half-to-three-quarter comparisons are within one
accelerator architecture, and comparisons between those pairs cross
architectures.
The one hardware effect we measured, a \ebTwoArchShift-point greedy shift
on another checkpoint (Appendix~\ref{app:disp}), does not bound hardware
effects on these sampled comparisons. The first twelve added
layers add \qnFiftyStep{} points of \heval{} \passat{10}, the next twelve
\qnSeventyfiveStep{}, the last twelve \qnDenseStep{}. The training-seed
spread of the full-density configuration at this budget is about
\qnSeedSpread{} points, wider than the whole range from half to full
density. We did not measure the uncertainty of the density contrasts. The
metrics do not agree on which densities differ (Table~\ref{tab:interface}).
GSM8K keeps rising with density, so full attachment scores \qnGsmGap{}
points higher than half density on arithmetic, and that comparison crosses
architectures. All four
arms use a uniform stride, and placement at matched density is measured
separately in Appendix~\ref{app:interface}.

\begin{table}[t]
\caption{\textbf{Interface ablations at a fixed tower.} Four
attach-density arms at 250\,M supervised tokens with the same corpus, seed
and harness. They differ in which tower layers the denoiser reads. The
25\,\% and 100\,\% arms were trained and scored on Hopper hosts, the 50\,\%
and 75\,\% arms on a Blackwell host (Appendix~\ref{app:repro}); seed spread
and unmeasured contrasts as in the text.}
\label{tab:interface}
\centering\small
\begin{tabular}{lcccc}
\toprule
attach density & \heval{} \passat{10} & GSM8K & \mmlupro{} & MBPP \\
\midrule
25\,\% (12/48) & \qnTwentyfiveHEten & \qnTwentyfiveGSM & \qnTwentyfiveMMLUP & \qnTwentyfiveMBPP \\
50\,\% (24/48) & \qnFiftyHEten & \qnFiftyGSM & \qnFiftyMMLUP & \qnFiftyMBPP \\
75\,\% (36/48) & \qnSeventyfiveHEten & \qnSeventyfiveGSM & \qnSeventyfiveMMLUP & \qnSeventyfiveMBPP \\
100\,\% (48/48) & \qnDenseHEten & \qnDenseGSM & \qnDenseMMLUP & \qnDenseMBPP \\
\bottomrule
\end{tabular}
\end{table}

\paragraph{Denoiser size.} Appendix~\ref{app:interface} measures a
smaller denoiser at a fixed tower and attach set. Two dense donors from the
same family, with \armbTrainable\,B and \armbTwoTrainable\,B trainable
parameters, are compared at the same 500\,M-token budget. The larger donor
is worse on every metric that separates the arms. At matched tokens per trainable
parameter (Appendix~\ref{app:interface}) it is better on \heval{} and worse
on \mmlupro{}, so a larger denoiser shows no benefit across the board. Its
learning-rate path also departed from the schedule
(Appendix~\ref{app:prereg}). The two donors also ran on different
accelerator architectures with different micro-batch sizes, so this
ablation is confounded three ways. We draw no size conclusion from it, and
a rerun with a matched schedule, host and micro-batch is the next step.

\paragraph{Limitations.} The dense comparison uses one seed per arm at 500\,M tokens and the main
MoE comparison uses 1\,B tokens. The dense frozen-tower model was still improving when stopped. We replicate the frozen-tower model across seeds but not the in-place
model. We did not run a frozen tower with the alignment loss; the trainable-tower
control (\S\ref{sec:theory}) is the nearest single-variable test of freezing,
at one seed and approximately one budget. The propositions address expressivity and gradient structure, the drift
and held-out measurements use one checkpoint each, and none addresses
optimisation dynamics or the empirical $500\times$ budget ratio. Batch-size-1 throughput remains below served AR, and all experiments use
post-RoPE key/value concatenation. We vary attach density at one budget and denoiser size at two
(\S\ref{sec:interface}), with one seed per arm. The density arms use a uniform stride, so they bound the effect of density
and say nothing about placement. The committed scores are single decode
draws. We measure their spread directly (\S\ref{sec:dispersion}).

\section{Conclusion}

Across the tested parents, the frozen-tower configuration retains
substantially more generation performance than the in-place configuration at
matched supervised-token budgets. Across masked-diffusion recipes on a
small parent and matched MoE and dense comparisons, in-place training
preserves most knowledge-benchmark performance and degrades generation on longer-output tasks. On the 30B MoE parent the frozen-tower configuration reaches \qttHEten{} on
\heval{} \passat{10} against \dfiveHEten{}, a factor of $\gtrsim11$. The frozen-tower model stays ahead of
the in-place model at every attach density we tested. At the same token
budget, a smaller denoiser sees more training tokens per trainable parameter
than a larger one. A token-matched comparison therefore cannot determine
whether a performance difference is caused by denoiser size or by the amount
of training each parameter receives. At 1\,B tokens, the frozen-tower model reaches \rndoneRetLo--\rndoneRetHi\% of the
500\,B-token
in-place reference score, while weight aliasing keeps resident weights
at the parent plus \towerOverheadGB\,GB.

Within the two-tower design, two models trained with the same recipe, one
with a frozen tower and one with a trainable tower, achieve similar observed
code scores, but the frozen-tower model scores \ttMMLUPdropAbs{} points
higher on \mmlupro{} in this comparison. The results therefore support
keeping the pretrained context representation fixed and available while
training the denoiser for parallel emission. The propositions show that the class contains a sampler
that reproduces the parent under a sequential schedule. They also show that
freezing removes one gradient term. Evaluation protocol is a separate
source of variation, the same dLLM score changing by up to $4.8\times$
across the tested protocols, so comparisons need a fixed task set,
parent-agreement controls and separate sampled and deterministic results.

\section{Related work}\label{sec:related}

\paragraph{\atod{} conversion.} Masked-diffusion conversion of AR checkpoints
spans DiffuGPT and DiffuLLaMA \citep{gong2025diffullama}, Dream
\citep{ye2025dream}, \rndone{} \citep{rnd1report2025}, and LLaDA2.0's 16B and
100B MoE conversions \citep{llada2025v2}, at budgets from under 200\,B to
500\,B tokens. Fast-dLLM~v2 \citep{wu2025fastdllm2} and block objectives of
the same family bring the cost to $\sim$1\,B tokens at 1.5--7B dense scale.
FLUID \citep{ma2026fluid} adapts dense models with a strictly causal
alignment and an entropy-driven denoising stride at a few billion tokens.
From-scratch diffusion language
models at scale include LLaDA \citep{nie2025llada} and LLaDA-MoE
\citep{zhu2025lladamoe}. None of this work reports a matched in-place against frozen-tower comparison
on one parent, corpus, budget and trainable set, a common-protocol
cross-model table, or the behaviour of expert routing under conversion.
Appendix~\ref{app:related} places the objectives and samplers we build on.

\paragraph{Two-tower conversion.} \citet{reda2026twotower} introduced the
frozen context tower, freezing a Mamba-hybrid MoE parent and training a
denoiser copy with layer-aligned cross-attention at \twotowerBudget{} tokens,
retaining \twotowerRetention\% of parent quality at
\twotowerSpeedup$\times$ throughput. Their parent is mostly recurrent and
exposes no causal attention mask to replace, so it admits no matched in-place
control. We adopt their geometry at $1/2100$ of the budget on a full-softmax
parent and supply the comparison and the analysis. With weight aliasing, resident weights total the parent plus
\towerOverheadGB\,GB, at the cost of a second forward
(\S\ref{sec:throughput}). In-place recipes such as Fast-dLLM~v2 avoid the
second forward.

\label{endmain}
\clearpage
\subsection*{AI use statement}

The authors used an LLM-based coding assistant for implementation and
experiment operation: writing and reviewing training, evaluation and analysis
code, operating runs on remote hosts, and drafting and checking prose in this
paper and in the internal results documents it cites. The assistant also
searched for related work and verified every reference. The propositions in
\S\ref{sec:theory} were formulated by the authors. The assistant checked the
proofs in Appendix~\ref{app:theory} and polished their writing. The assistant
drafted the documents that fixed bands before each run, calibration bars and results readings from
the authors' stated designs, proposed candidate experiments and readings, and
produced first drafts of the analyses in the appendices. Every band, stopping
rule and reading rule was approved by the authors before the corresponding
run. The outcomes were not fixed in advance. Four runs departed from their
documents. Appendix~\ref{app:prereg} records each departure and when it was
written down, including one check that was adopted after its cell had been
measured. Every interpretation in this paper was reviewed against
the committed evidence by the authors. And every number reported here was
produced by the committed code on the frozen harness and verified against the
committed artefacts. The assistant is not an author and made no research
contribution that the authors did not direct and check.

\subsection*{Ethics statement}

This work trains and evaluates language models on public benchmarks and a
public pretraining mixture. It involves no human subjects and collects no
data. Converted models inherit the capabilities and
limitations of their parents and will be released for research use.

\subsection*{Reproducibility statement}

For every training run the numerical bands, stopping rules and reading
rules were fixed in a signed document before the run launched, and every
arm was scored on one frozen evaluation harness. Analyses added after
review, such as the within-task length grouping in
Appendix~\ref{app:length}, are marked as exploratory where they appear. Appendix~\ref{app:prereg} lists the bands the paper reads against and the
four departures from them. Appendix~\ref{app:harness}
specifies the task suites and sampling parameters;
Appendix~\ref{app:disp} gives the per-seed decode tables behind
\S\ref{sec:dispersion}; Appendix~\ref{app:repro} inventories every arm: the
configuration of record and seed, corpus, training host, and which
artefacts the archived record holds. Configurations are committed once per run and never edited
in place. Committed numbers are never revised; new measurements are reported
as new rows, and where a later result changes an interpretation we change the
wording and leave the number as measured. The configuration files, corpus
manifest, evaluation specification and harness revision will be released
with the code.

\bibliographystyle{iclr2027_conference}
\bibliography{axi_references}

\appendix

\section{Proofs and the toy model}\label{app:theory}

\paragraph{Proof of Proposition~\ref{prop:witness}.}
Index absolute positions from 1. The committed prefix $c$ occupies positions
$1,\dots,n$ and is held by the tower. The current block occupies positions
$n{+}1,\dots,n{+}S$. The canvas has $S{+}1$ slots: slot $0$ is the anchor and it
holds the last committed token $c_n$ at position $n$, and slot $q\ge1$ holds
block position $n{+}q$. At round $i$ ($1\le i\le S$), slots $1,\dots,i{-}1$
hold the committed block tokens $x_{B,<i}$ and slots $i,\dots,S$ hold mask
tokens. Define $\phi^\dagger$ by copying every module from $\theta^*$: the
experts and embeddings are bit-identical by construction, and the attention,
router, normalisation layers and output head take their parent values. The
visible set of a canvas query at slot $q$ is
$V_q=\{1,\dots,n\}\cup\{\text{slots } r: 1\le r\le\min(q,i{-}1)\}$, the
tower's keys for the prefix together with the committed canvas slots at or
before the query; masked slots and later committed slots are suppressed, and
the anchor's own key is suppressed because position $n$ is already exported
by the tower. The emission for block position $i$ is read at slot $i{-}1$,
the slot holding the token at position $n{+}i{-}1$ (the anchor when $i=1$).

We show by induction over layers that, for every visible slot $r\le i{-}1$
(the anchor included), the hidden state at slot $r$ equals the parent's
hidden state at position $n{+}r$ for the sequence $(c,x_{B,<i})$. At layer
$0$ the input at slot $r$ is the embedding of the committed token at position
$n{+}r$, as in the parent, and every layer rotates its queries and keys at
that same absolute position. For the
inductive step, assume equality through layer $l{-}1$ at every visible slot.
At layer $l$ the query at slot $r$ attends over
$[\mathrm{KV}_{\theta^*}(c)\Vert\mathrm{KV}(\text{slots }1,\dots,r)]$. The
tower's keys and values are the parent's for positions $1,\dots,n$, since the
tower is the parent run causally; the canvas keys and values at slots
$1,\dots,r$ are the parent's for positions $n{+}1,\dots,n{+}r$ by the
induction hypothesis, since keys are rotated at their absolute positions
and values are projected from the same hidden states. This is exactly the set the parent's causal attention
sees at position $n{+}r$, and grouped-query repetition is identical, so the
attention output at slot $r$ equals the parent's at position $n{+}r$. The
identical router, experts and normalisation layers close the step. The output
head at slot $i{-}1$ therefore yields $\pi^*(\cdot\mid c,x_{B,<i})$, and the
one-position-per-round left-to-right schedule factorises the block exactly.
When the block completes, its $S$ tokens are committed to the tower's prefix
and the next block's anchor holds $x_{B,S}$ at its absolute position, so the
argument repeats across block boundaries. \hfill$\square$

The equivalence is exact under the visibility assumption. The agreement of
the implemented witness with the parent in finite precision
(Appendix~\ref{app:prereg}, row 3) is a separate check on the implementation.

\begin{proposition}[Exact gradient decoupling]\label{prop:decouple}
Let $L(\psi)=\mathbb{E}\,\ell(q_\psi(z,\mathrm{KV}_\psi(c)))$. For the
in-place parameterisation, where the context states depend on the trained
weights $W$, the gradient is
$\nabla_W L=G_{\mathrm{emit}}(W)+J_W^{\top}G_{\mathrm{subst}}(W)$. For the
frozen-tower parameterisation, the context states are constant and the
gradient is $G_{\mathrm{emit}}$ alone; at the parent initialisation
$W=\theta^*$, where the context states coincide, this term is the same in
both arms. The parameterisations therefore differ by exactly one term,
$J_W^{\top}G_{\mathrm{subst}}$, which is not constrained to preserve the
witness in Proposition~\ref{prop:witness}.
\end{proposition}

\paragraph{Proof of Proposition~\ref{prop:decouple}.}
By the chain rule, for
$L(W)=\mathbb{E}\,\ell(q_W(z,\mathrm{KV}_W(c)))$ with $W$ appearing in both
arguments,
$\nabla_W L=\mathbb{E}\big[(\partial_W q)^{\top}\nabla_q\ell\big]
+\mathbb{E}\big[(\partial_W\mathrm{KV}_W(c))^{\top}
(\partial_{\mathrm{KV}}q)^{\top}\nabla_q\ell\big]$,
where $\partial_W q$ is the Jacobian of $q$ with respect to $W$ at fixed
context states, $\partial_{\mathrm{KV}}q$ is its Jacobian with respect to
the context states, and $\partial_W\mathrm{KV}_W(c)$ is $J_W$. The first
expectation is $G_{\mathrm{emit}}$, and $J_W^{\top}G_{\mathrm{subst}}$ is
shorthand for the second. The frozen-tower model computes
$\mathrm{KV}_{\theta^*}(c)$ with constant $\theta^*$, so the second term
vanishes. At $W=\theta^*$, where the context states coincide, its first term is the same
$G_{\mathrm{emit}}$ as in the in-place model. \hfill$\square$

\paragraph{The two-block toy model.} Consider data $(x,y)$ with
$y=A_2^*A_1^*x+\xi$, substrate $A_1\in\mathbb{R}^{k\times d}$, emission map
$A_2\in\mathbb{R}^{m\times k}$, and squared loss. With $A_1\equiv A_1^*$, the objective is a least-squares problem in $A_2$
alone. Under joint training, $\nabla_{A_1}L$ contains terms of the form
$A_2^{\top}Rx^{\top}$, where $R$ is the residual. When the emission map is
initially misaligned with the new task, these terms are generically nonzero,
so joint training can move $A_1$ away from $A_1^*$ while $A_2$ is still
aligning. This is the mechanism \citet{kumar2022finetuning} formalise for
fine-tuning relative to linear probing; the model here is a qualitative
illustration and derives no rate.

The toy model is consistent with three observations: flat low-budget
in-place trajectories, insensitivity to trainable-parameter count, and a gap
that increases with the required degree of parallelism. It is a qualitative
illustration and does not account for the observed transformer gap.

\paragraph{Components of the frozen-tower intervention.} The frozen-tower
intervention combines three changes: a fixed context representation, an
additional read path, and gradient isolation. Proposition~\ref{prop:decouple}
formalises the last of these. A sequence of measured interventions provides a
partial test of context anchoring. On the small parent, the unanchored
in-place model scores \ponHEten{} on \heval{} \passat{10}. Adding layer-wise
representation alignment as a loss raises the score to \ptwoHEten{}, with a
1.8\% increase in training loss. On the 30B parent, architectural anchoring
with the frozen tower reaches \qttHEten{}, compared with \dfiveHEten{} for
the matched in-place model. These comparisons do not isolate context
anchoring: the loss-based and architectural interventions use different
parent scales. The trainable-tower control below more directly tests
freezing within the two-tower design and gives different results on
\heval{} and \mmlupro{}. The propositions establish realisability and gradient structure, not
optimisation dynamics. The $11.6\times$ pair difference and the
$500\times$ budget ratio are measured rather than derived.

\paragraph{The trainable-tower control: record.} Table~\ref{tab:control}
reads the control against the committed frozen seed-1 checkpoint. The
configuration is the frozen recipe with four changes: the tower is
trainable, training stops at 500\,M tokens, a checkpoint is saved at step
2{,}480, and the tower is held resident in memory. The last of these affects
where the tower's weights live and not what is computed. With the flag off,
the acceptance record found identical outputs under a deterministic kernel.
Corpus, data order,
RNG seed 42, corruption seed, loss, block schedule, learning rate (1e-4,
2{,}000-step warm-up, cosine over the 1\,B run) and decoding are shared.

Two differences remain. The control's final checkpoint is at 5{,}059 steps
and 500{,}079{,}173 supervised tokens, and the frozen comparator is the
lineage's step-4{,}960 save at about 490\,M tokens, so the two sit 99 steps
apart on the shared schedule. The control was also scored on an H200 and the
comparators on a GH200. Both are Hopper hosts, and the decode path was
checked token-identical. The second frozen seed ran 5{,}050 steps on
499{,}236{,}606 tokens, close to the control's budget, and scores
\qttSTwoHEtenB{} and \qttSTwoMMLUP{}. Training the tower raises the
trainable count from 0.92\,B to 1.84\,B. Every evaluation decoded through
the trained tower: 427 of the 433 tower tensors differ from the parent, and
the six that do not are never read.

\paragraph{Uncertainty.} The committed frozen \heval{} draw (\qttHEtenB{})
has no per-problem file, so the paired intervals use the same checkpoint's
three re-decodes from Appendix~\ref{app:disp}. Against the re-decode at
decode seed 1234 (64.15) the difference is $-0.33$ [$-6.13$, $+5.26$].
Against seeds 2345 (63.57) and 3456 (63.87) it is $+0.26$ [$-5.64$,
$+6.41$] and $-0.04$ [$-5.42$, $+5.37$]. On \hevalplus{} the three read
$-1.43$, $+0.32$ and $-0.58$, with intervals of similar width. Each
interval comes from 10{,}000 resamples of the 164 problems, with
\passat{10} recomputed from the per-problem counts. They describe movement
across problems for these checkpoints and generations. They do not measure
training-seed variability, and all three use the same control draw.

The frozen 500\,M \mmlupro{} score has no per-item file, so no paired
interval can be formed for that comparison. The interval in
Table~\ref{tab:control} treats the two scores as independent binomials,
204/500 against 271/500, with a standard error of 3.13 points and a normal
95\,\% width. It ignores the dependence that comes from scoring both models
on the same questions. A paired interval exists only against the frozen
1\,B final (54.6, 273/500), which gives $-13.8$ [$-18.0$, $-9.4$] over item
resamples at a different budget.

At 250\,M the control reads \ttMMLUPT{} against \ttMMLUPTwoFifty{} on
\mmlupro{} and \ttHEtenT{} against \qttHEtenA{} on \heval{}. Every other
metric moves by under four points: GSM8K \ttGSM{} against \qttGSMfiveH{},
MBPP \ttMBPP{} against \qttMBPPfiveH{}, and \hevalplus{} \ttHEplusTen{}
against \qttHEplusfiveH{}. The bands were 61.0--67.0 on \heval{} for a
matched reading and below 51.2 on \mmlupro{} for a loss. Both were met. The
registered conflict rule covers co-primaries that point in opposite
directions, and a matched score points in no direction, so it was not
invoked. A band is a decision rule and not an equivalence test. Training the
tower costs \ttPeakDelta{}\,GiB of peak memory and makes a step
$\ttStepFactor\times$ slower. The control is one seed at one budget, and
\mmlupro{} is one benchmark.

\paragraph{Scope.} The propositions establish realisability and the exact
one-term gradient difference between the two parameterisations. The toy model
illustrates the mechanism qualitatively. These results do not predict
training-budget constants for transformers; the $500\times$ ratio is measured
rather than derived.

\section{Mechanism measurements on the matched pair}\label{app:mech}

The two configurations of \S\ref{sec:pair} share parent, corpus and data
order, supervised-token budget, trainable parameter set, optimiser and schedule,
denoising-objective family and harness, and differ in four things. \dfive{} trains under
a representation-alignment loss to the frozen parent (cosine, $\lambda{=}10$,
all layers), with a canvas block that ramps from 32 to 64 tokens at step
1000. Its corruption masks each position on an independent draw, at a rate
that depends on the position. \qtt{} has no alignment loss, because its
tower is the parent and never moves. Its canvas block stays at 32
throughout. Its corruption masks exactly $k$ positions per block, drawn by
Gumbel top-$k$ with a weighting towards later positions. The in-place block
objective processes $4\times$ as many tokens per supervised token, so on the
same hardware the frozen-tower model trains about $3.6\times$ faster per
supervised token, and the 500\,M-token replication run took 27.4\,h on one
GPU. Each architecture decodes under its own backend on one harness
(Appendix~\ref{app:harness}). The comparison therefore evaluates the two
configurations; the trainable-tower comparison described in
Appendix~\ref{app:theory} tests freezing within the two-tower design, with
the checkpoint-budget and host differences recorded there.

Both measurements below use the committed 1\,B checkpoints of the pair and
the parent, with the bands fixed before they ran. No number in
\S\ref{sec:pair} or \S\ref{sec:theory} changed. The held-out sequences come
from the shared corpus under the seed-42 permutation, from index 700{,}000
onward. Both runs consumed 647{,}040 sequences, so the held-out slice begins
past everything either of them saw.

\subsection{Context drift under in-place conversion}\label{app:drift}

Proposition~\ref{prop:decouple} says the in-place gradient carries a term
that can move $\mathrm{KV}_W(c)$ off $\mathrm{KV}_{\theta^*}(c)$. This
measures whether the maps moved. On 2{,}000 held-out sequences, \dfive{} and the
parent are run causally over the same clean tokens, and at every supervised
position the per-layer keys (after \texttt{k\_norm} and RoPE at absolute
positions) and values are compared by relative Frobenius distance, with the
cosine of the residual stream as a second view. The frozen arm's tower is the
parent, so its distance is zero by construction. This comparison and
parent-against-parent were run as sanity checks and returned exactly zero.

\textbf{Reading: DRIFT-PRESENT.} The prespecified band was a mean relative
distance of at least 0.05 in a quarter of the layers or more. Measured: mean
relative K distance \driftKmean{} and V distance \driftVmean{}, above 0.05
in all 48 layers, with a maximum of \driftMax{} (layer 36, V);
residual-stream cosine to the parent \driftCosMean{} on average and
\driftCosMin{} at its lowest (Figure~\ref{fig:drift}). The profile rises with
depth, from 0.18/0.29 at layer 0 to about 0.38/0.55 over the last twelve
layers: the maps moved everywhere, while the residual stream kept its
direction to within a few degrees. No prediction about where drift concentrates was set down in advance, so the profile confirms none. Only the 1\,B
checkpoint exists (the 250\,M--750\,M in-place checkpoints were not
archived), so no trajectory is claimed.

\begin{figure}[H]
\centering
\begin{tikzpicture}
\begin{axis}[width=6.3cm, height=5.2cm, xmin=-1, xmax=48, ymin=0, ymax=0.7,
  xlabel={layer}, ylabel={relative distance to the parent},
  title={\footnotesize \dfive{} key/value maps, 2{,}000 held-out sequences},
  tick label style={font=\footnotesize}, label style={font=\footnotesize}, grid=major, major grid style={black!12},
  legend style={at={(0.03,0.97)}, anchor=north west, font=\scriptsize, draw=black!30}]
\addplot[red!70!black, mark=*, mark size=1.2pt, thick] coordinates {(0,0.181) (1,0.294) (2,0.293) (3,0.451) (4,0.314) (5,0.297) (6,0.307) (7,0.344) (8,0.313) (9,0.277) (10,0.287) (11,0.289) (12,0.279) (13,0.342) (14,0.324) (15,0.318) (16,0.308) (17,0.328) (18,0.309) (19,0.285) (20,0.284) (21,0.276) (22,0.319) (23,0.347) (24,0.277) (25,0.352) (26,0.336) (27,0.347) (28,0.344) (29,0.317) (30,0.369) (31,0.300) (32,0.305) (33,0.301) (34,0.379) (35,0.363) (36,0.288) (37,0.367) (38,0.348) (39,0.372) (40,0.359) (41,0.329) (42,0.385) (43,0.383) (44,0.416) (45,0.399) (46,0.355) (47,0.379)};
\addlegendentry{K}
\addplot[blue!70!black, mark=square*, mark size=1.2pt, thick] coordinates {(0,0.288) (1,0.265) (2,0.372) (3,0.297) (4,0.372) (5,0.392) (6,0.369) (7,0.385) (8,0.314) (9,0.351) (10,0.352) (11,0.375) (12,0.405) (13,0.422) (14,0.399) (15,0.400) (16,0.408) (17,0.400) (18,0.423) (19,0.437) (20,0.389) (21,0.421) (22,0.453) (23,0.483) (24,0.508) (25,0.526) (26,0.489) (27,0.461) (28,0.466) (29,0.468) (30,0.490) (31,0.502) (32,0.466) (33,0.497) (34,0.585) (35,0.600) (36,0.601) (37,0.590) (38,0.510) (39,0.487) (40,0.495) (41,0.514) (42,0.408) (43,0.452) (44,0.484) (45,0.422) (46,0.463) (47,0.548)};
\addlegendentry{V}
\addplot[black!40, dashed] coordinates {(-1,0.05) (48,0.05)};
\addlegendentry{band (0.05)}
\end{axis}
\end{tikzpicture}\hfill
\begin{tikzpicture}
\begin{axis}[width=6.3cm, height=5.2cm, xmin=-1, xmax=48, ymin=0.9, ymax=1.0,
  xlabel={layer}, ylabel={residual-stream cosine},
  title={\footnotesize same sequences, same positions},
  tick label style={font=\footnotesize}, label style={font=\footnotesize}, grid=major, major grid style={black!12}]
\addplot[black!70, mark=*, mark size=1.2pt, thick] coordinates {(0,0.9904) (1,0.9847) (2,0.9831) (3,0.9811) (4,0.9795) (5,0.9795) (6,0.9793) (7,0.9799) (8,0.9801) (9,0.9799) (10,0.9798) (11,0.9788) (12,0.9774) (13,0.9763) (14,0.9760) (15,0.9750) (16,0.9749) (17,0.9743) (18,0.9729) (19,0.9726) (20,0.9716) (21,0.9708) (22,0.9695) (23,0.9681) (24,0.9660) (25,0.9653) (26,0.9658) (27,0.9646) (28,0.9662) (29,0.9647) (30,0.9638) (31,0.9637) (32,0.9619) (33,0.9596) (34,0.9584) (35,0.9563) (36,0.9542) (37,0.9519) (38,0.9527) (39,0.9487) (40,0.9493) (41,0.9463) (42,0.9464) (43,0.9469) (44,0.9496) (45,0.9513) (46,0.9548) (47,0.9579)};
\end{axis}
\end{tikzpicture}
\caption{\textbf{Context drift of the in-place arm after 1\,B tokens.} Left:
per-layer relative Frobenius distance of \dfive{}'s keys and values from the
parent's on the same held-out tokens; the frozen arm sits at zero by
construction. Right: cosine between the two residual streams.}
\label{fig:drift}
\end{figure}

\subsection{Held-out objective of the pair}\label{app:elbo}

We scored each arm under its own training objective, on the same 25{,}000
held-out sequences with the same corruption seed (777). The two kernels
differ, so the arms do not see the same masks. \dfive{} was scored with
independent per-position masking at block 64, and \qtt{} with exactly-$k$
masking at block 32. Each score is the diffusion term alone, with no
representation alignment and no router auxiliaries. We scored one batch
twice and got the same loss to the digit.

\textbf{Reading: between the bands, reported as measured.} Two bands were
set in advance. OBJECTIVE-MATCHED was an in-place loss within 0.10 nats
(natural-log units per scored token) of the frozen arm's, and
OBJECTIVE-WORSE was a loss 0.30 nats or more above it. The measured losses are
\elboInplaceFull{} nats for \dfive{} and \elboFrozenFull{} for \qtt{}
(Table~\ref{tab:elbo}). The in-place arm is \elboGap{} nats \emph{below}
the frozen arm. That falls outside the 0.10 band, on the side no band
named, so the result is reported as measured.

The two losses are not on a common scale. The objectives differ in
corruption kernel, mask geometry and block size, so the losses do not
measure how far each arm has converged relative to the other. The
OBJECTIVE-MATCHED reading would not have established that either. Both arms
reduce their own denoising loss by about twelve nats from initialisation,
while their generation performance differs sharply.

The last two rows of the table were added after the readings. They score
the parent initialisation under each arm's objective. The arms start at
\elboParentDfive{} and \elboParentQtt{} nats and reduce their own objectives
by \elboImprInplace{} and \elboImprFrozen{} (paired over sequences, sd
\elboImprInplaceSd{} and \elboImprFrozenSd{}). Each reduction is in that
arm's own units and is reported as such.

\begin{table}[H]
\caption{\textbf{Held-out objective of the pair.} 25{,}000 held-out
sequences, corruption seed 777, each arm under its own objective, in nats
per scored token. The parent rows score the untrained initialisation of
each arm under that arm's objective. Two sequences in the frozen arm's draw
carried no scored token. They enter the token-weighted mean with zero
weight, and the paired statistics use the remaining 24{,}998. The last two
rows were added after the readings.}
\label{tab:elbo}
\centering\small
\resizebox{\linewidth}{!}{%
\begin{tabular}{lcc}
\toprule
 & \dfive{} (in-place) & \qtt{} (frozen tower) \\
\midrule
mean loss, token-weighted & \elboInplaceFull{} & \elboFrozenFull{} \\
mean of per-sequence losses & \elboInplaceSeq{} & \elboFrozenSeq{} \\
paired difference, frozen $-$ in-place & \multicolumn{2}{c}{$+\elboPairedMean{}$ (sd \elboPairedSd{}), in-place worse on \elboInplaceWorseFrac{}\% of sequences} \\
parent initialisation, same objective (unregistered) & \elboParentDfive{} & \elboParentQtt{} \\
improvement from initialisation & \elboImprInplace{} & \elboImprFrozen{} \\
\bottomrule
\end{tabular}}
\end{table}

\subsection{The witness and the K-sweep}\label{app:witness}

Proposition~\ref{prop:witness} says the frozen-tower class contains a
sampler that reproduces the parent. We call that sampler the witness,
because running it shows the class really does contain one. We built it and
ran it on one Hopper host (Appendix~\ref{app:repro}). The original bands were fixed before any cell ran, and the later
criterion change is recorded in Appendix~\ref{app:prereg}. The construction starts from the untrained
frozen-tower initialisation, so the parent is both denoiser and tower, and
it adds two options to the sampler. The first option is the acceptance
order. It commits the leftmost masked position each round, in place of the
most confident one. The second option is masked-key suppression. It hides
from each query the keys of masked positions, of committed positions after
it, and of the anchor slot. These are the two options the checks below turn
off and on. One more setting is fixed by the model and is never toggled.
Each model reads its next token where it was trained to read it, so the
parent reads at the last committed position and the trained final at the
position itself. The witness also stops at the end-of-text token.

Two checks were registered. First, with both options off, the trained
frozen final reproduces its committed greedy \witParity{} exactly. All 164
generations are byte-identical to the committed run, on a different box,
driver and torch build. Second, with both options on, the untrained witness
scores \witKone{} greedy \passat{1} on \heval{}. That equals the parent's
committed own-decode score, \parentHEgreedyParity{}, to the digit.

The parent has three greedy scores in this appendix, and they differ.
Served by vLLM it scores $53.66$, which is what the band was written
against. On the frozen harness's own decode path it scores
\parentHEgreedyParity{}, measured and committed before the band was set.
Re-decoded on the witness's own box with the same script it scores
\witParentSameBox{}. The witness matches the own-decode score to the digit,
falls outside the registered band of $53.66\pm0.5$, and sits
\witScoreGap{} points above the same-box score. After the first measurement
the reference was corrected to the own-decode score, and that correction is
reported as a deviation (Appendix~\ref{app:prereg}).

Token identity was replaced by an agreement criterion before any later cell
ran. The witness's committed tokens agree with the parent's teacher-forced
argmax on \witAgree\% of 22{,}632 positions. The parent's own free-run
decode agrees with its argmax on \witParentAgree\%, so two runs of this
parent do not agree token for token in bf16, and the witness clears the
parent's own level. The proposition establishes exact equivalence under its
assumptions. The implementation checks show close numerical agreement.

Under sampling ($n{=}20$, temperature 0.8, top-$p$ 0.95) the witness
reaches \witKoneSampled{} \passat{10} against the parent's \parentHEten{}.
The registered check was that this score fall inside the parent's
paired-bootstrap interval around \parentHEten{}. That interval could not be
formed, because the parent's cell is a committed summary with no
per-problem file. The check reported instead is that the parent's value lies inside the
witness's own 95\,\% problem-resample interval $[\witIntLo, \witIntHi]$
(Appendix~\ref{app:prereg}).

\paragraph{The sweep.} Committing $K$ positions per round with zero training
gives Table~\ref{tab:witness}. The schedule is exact at every $K$: $32/K$
rounds per block and $K$ tokens per forward. Only $K{=}1$ scores. Greedy \passat{1} is zero at every tested $K$ above
1; sampled \passat{10} is also zero at $K{=}4$, the only other $K$ sampled.
The second position in a round is
read from a slot whose predecessor has not been committed yet, and the
untrained parent has nothing to put there. Commit entropy rises with $K$,
from \witEntKone{} nats at $K{=}1$ to \witEntKmax{} at $K{=}32$.

\begin{table}[H]
\caption{\textbf{The untrained witness against tokens committed per
forward.} Greedy \passat{1} on \heval{}, masked keys suppressed, $S{=}32$,
164 problems per cell. Sampled \passat{10} ($n{=}20$) was measured at
$K{=}1$ (\witKoneSampled{}) and $K{=}4$ (\witKfourSampled{}). No band was
placed on this curve.}
\label{tab:witness}
\centering\small
\begin{tabular}{lcccc}
\toprule
$K$ & greedy \passat{1} & rounds per block & tokens per forward & commit entropy (nats) \\
\midrule
1 & \witKone{} & 32 & 1 & \witEntKone{} \\
2 & \witKtwo{} & 16 & 2 & 1.53 \\
4 & 0.00 & 8 & 4 & 1.65 \\
8 & 0.00 & 4 & 8 & 1.66 \\
16 & 0.00 & 2 & 16 & 1.92 \\
32 & 0.00 & 1 & 32 & \witEntKmax{} \\
\bottomrule
\end{tabular}
\end{table}

\paragraph{Applying the witness decoding rules to the trained model.} Before
the run we set down one quantity to report at $K{=}4$, the gap between the
untrained witness and the trained model's \qttHEten{}. The two numbers do
not separate what the architecture gives from what the training gives. One
model is trained and the other is not, and they decode under different
rules. The trained model's own sampler commits about \qttTokPerFwd{} tokens
per forward on average, while the witness rule commits exactly four.
Under the witness rules the trained final scores \witTrainedKfour{} at
$K{=}4$, greedy and sampled. Toggling the two options one at a time on the
trained final (greedy, $K{=}4$) separates them. Masked-key suppression alone
takes its \witParity{} to \witSupOnly{}. The left-to-right order alone takes
it to \witLrOnly{}. Both together take it to zero. The toggles show
sensitivity to the tested decoding changes and do not establish why one
costs more than the other. The trained model was trained with the keys that
suppression hides visible to every query. The tested witness decoding
settings do not preserve the trained checkpoint's generation performance, so
the existence result of Proposition~\ref{prop:witness} does not establish
compatibility with the sampler used after denoising training.

\section{The dense-parent matched comparison}\label{app:dense}

The dense-parent arms are read in \S\ref{sec:pair}, which states every
number below in prose; the table is reproduced here for reference.

\begin{table}[H]
\caption{\textbf{Dense-parent matched comparison.} Each arm uses
\dpsBudget{} supervised tokens with the same corpus, seed, and evaluation
harness; each is evaluated with its native decoder. The ``\%'' column gives
the frozen-tower score as a percentage of the parent score. The
prespecified replication criterion is met:
$r_s=\dpsRatio{}\geq0.45$ and $r_i=\dpiRatio{}\leq\dpiCeil{}$.}
\label{tab:dense}
\centering\small
\resizebox{\linewidth}{!}{%
\begin{tabular}{lcccc}
\toprule
 & \dparent{} (parent) & \dps{} & \% & \dpi{} (in-place) \\
\midrule
\heval{} \passat{10} & \dparentHEten & \dpsHEten & \dpsRetPct & \dpiHEten \\
\heval{} greedy \passat{1} & \dparentHEgreedy & \dpsHEgreedy & 63.3 & \dpiHEgreedy \\
GSM8K & \dparentGSM & \dpsGSM & 85.4 & \dpiGSM \\
\mmlupro{} & \dparentMMLUP & \dpsMMLUP & 97.9 & \dpiMMLUP \\
MBPP & \dparentMBPP & \dpsMBPP & 60.7 & \dpiMBPP \\
\midrule
BoolQ & \dparentBoolq & \dpsBoolq & 99.8 & \dpiBoolq \\
SQuAD-short EM & \dparentSquad & \dpsSquad & 93.6 & \dpiSquad \\
JSON validity & \dparentJSONvalid & \dpsJSONvalid & 98.4 & \dpiJSONvalid \\
\bottomrule
\end{tabular}}
\end{table}

\section{Output length and retention}\label{app:length}

The main text orders tasks by output length and reads degradation along that
order. This appendix measures the axis with a proxy which is the parent's generated
length: the mean number of tokens the \emph{parent} generates for each task
under the frozen harness, counted with the parent's tokenizer. For the two multiple-choice tasks
the answer is one token and the harness cap is not binding. GSM8K and MBPP hit
the 256-token generation cap in every parent sample, so their length is
censored at 256: the observed generated length is censored at the cap. We
did not rerun these tasks with a higher cap, so the axis is bounded at 256
for them and their true stopping length is unknown. The
association
between length and retention is read across tasks. Therefore it is not a causal
account of any one task. Retention is defined by the arm's score divided by the parent's on the
same task.

\begin{figure}[H]
\centering
\begin{tikzpicture}
\begin{axis}[width=6.9cm, height=5.6cm, xmode=log, xmin=0.7, xmax=420, ymin=-0.05, ymax=1.7,
  xlabel={parent's mean generated length (tokens, log scale)}, ylabel={retention (arm / parent)},
  title={\footnotesize 30B MoE pair, 1\,B tokens}, xtick={1,10,100,256}, xticklabels={1,10,100,256},
  tick label style={font=\footnotesize}, label style={font=\footnotesize}, grid=major, major grid style={black!12},
  legend style={at={(0.03,0.03)}, anchor=south west, font=\scriptsize, draw=black!30}]
\addplot[only marks, red!70!black, mark=*, mark size=2pt] coordinates {(1.0,0.82) (1.1,0.99) (7.7,1.22) (18.3,0.32) (138.6,0.07) (256,0.09) (256,0.01)};
\addlegendentry{\dfive{} (in-place)}
\addplot[only marks, blue!70!black, mark=square*, mark size=2pt] coordinates {(1.0,0.99) (1.1,1.00) (7.7,1.61) (18.3,0.98) (138.6,0.80) (256,0.95) (256,0.72)};
\addlegendentry{\qtt{} (frozen tower)}
\addplot[black!40, dotted, thick] coordinates {(256,-0.05) (256,1.7)};
\end{axis}
\end{tikzpicture}\hfill
\begin{tikzpicture}
\begin{axis}[width=6.9cm, height=5.6cm, xmode=log, xmin=0.7, xmax=420, ymin=-0.05, ymax=1.7,
  xlabel={parent's mean generated length (tokens, log scale)},
  title={\footnotesize 4B dense pair, 500\,M tokens}, xtick={1,10,100,256}, xticklabels={1,10,100,256},
  tick label style={font=\footnotesize}, label style={font=\footnotesize}, grid=major, major grid style={black!12},
  legend style={at={(0.03,0.03)}, anchor=south west, font=\scriptsize, draw=black!30}]
\addplot[only marks, red!70!black, mark=*, mark size=2pt] coordinates {(1.0,0.69) (1.1,0.92) (7.7,0.53) (18.3,0.10) (138.6,0.08) (256,0.07) (256,0.00)};
\addlegendentry{\dpi{} (in-place)}
\addplot[only marks, blue!70!black, mark=square*, mark size=2pt] coordinates {(1.0,0.98) (1.1,1.00) (7.7,0.94) (18.3,0.98) (138.6,0.61) (256,0.85) (256,0.61)};
\addlegendentry{\dps{} (frozen tower)}
\addplot[black!40, dotted, thick] coordinates {(256,-0.05) (256,1.7)};
\end{axis}
\end{tikzpicture}
\caption{\textbf{Retention against the parent's generated length.} Each
point is one task (Table~\ref{tab:length}); the dotted line is the 256-token
generation cap, at which generated length is censored. The trend is an
association across tasks. Both in-place arms fall with generated length, from 0.69--0.99
at one token to at most 0.1 at 139 tokens and beyond; both frozen-tower
arms stay near the parent until the longest tasks. SQuAD-short sits above 1
for the formatting reason given in \S\ref{sec:pair}.}
\label{fig:length}
\end{figure}

\begin{table}[H]
\caption{\textbf{The points of Figure~\ref{fig:length}.} Length is the
parent's mean generated tokens; $256^{+}$ marks the cap. Retention is
arm over parent on the score the task is read on (JSON validity for
CoNLL-JSON).}
\label{tab:length}
\centering\small
\begin{tabular}{lrcccc}
\toprule
 & & \multicolumn{2}{c}{30B MoE pair} & \multicolumn{2}{c}{4B dense pair} \\
task & length & in-place & frozen & in-place & frozen \\
\midrule
\mmlupro{} & 1 & 0.82 & 0.99 & 0.69 & 0.98 \\
BoolQ & 1.1 & 0.99 & 1.00 & 0.92 & 1.00 \\
SQuAD-short EM & 7.7 & 1.22 & 1.61 & 0.53 & 0.94 \\
CoNLL-JSON validity & 18.3 & 0.32 & 0.98 & 0.10 & 0.98 \\
\heval{} \passat{10} & 138.6 & 0.07 & 0.80 & 0.08 & 0.61 \\
GSM8K & $256^{+}$ & 0.09 & 0.95 & 0.07 & 0.85 \\
MBPP & $256^{+}$ & 0.01 & 0.72 & 0.00 & 0.61 \\
\bottomrule
\end{tabular}
\end{table}

\paragraph{Within one task.} To read the association without crossing
tasks, we split the 164 \heval{} problems into thirds by the length of the
canonical solution and score each arm per third, using the mean of its
three re-decoded seeds from Appendix~\ref{app:disp} (both arms on one
Blackwell host). Table~\ref{tab:withintask} gives the result. The in-place
arm solves problems almost only in the shortest third. The frozen-tower arm
falls with length and stays far above it in every third. This check was
added after review and was not prespecified. The parent has no per-problem
file, so retention per third is not formed. Canonical-solution length may
also correlate with problem difficulty, so this analysis does not isolate a
causal effect of output length. Solution length is the number of lines of
the canonical solution after trimming, blank and comment lines included;
problems are ordered by that count with ties broken by task id and split
into thirds, so a problem with three or six lines can fall on either side
of a boundary. No intervals were computed. The script and the group
membership will be released with the code.

\begin{table}[H]
\caption{\textbf{\heval{} \passat{10} by canonical-solution length}, thirds
of 55, 55 and 54 problems with median lengths of 1, 5 and 8 lines; each
arm's per-problem estimate is the mean of its three decode seeds.}
\label{tab:withintask}
\centering\small
\begin{tabular}{lcc}
\toprule
third by solution length & frozen tower (\qtt) & in-place (\dfive) \\
\midrule
shortest third & 81.65 & 13.43 \\
middle third & 70.28 & 0.00 \\
longest third & 50.18 & 0.31 \\
\bottomrule
\end{tabular}
\end{table}

The figure leaves two things open. CoNLL-JSON is the structured task and
also a mid-length one, and it sits lowest for its length in both families.
These tasks therefore cannot separate a length effect from structured
output failing before free text does. The uniform-state arm of
Appendix~\ref{app:negatives} speaks to the structured-output reading.

The in-place arms are not failing because they run out of room. On GSM8K
and MBPP they generate 190 and 152 tokens on average, while the frozen arms
use the full 256. Part of the failure at length is stopping early.

Fixed-length canvases are a known weakness of block-wise diffusion. Block
diffusion \citep{arriola2025bd3lm} allows variable-length generation,
DreamOn \citep{wu2026dreamon} adds length-control states so a canvas can
grow or shrink, and FLUID \citep{ma2026fluid} adapts the denoising stride
to local information density. Our length result is about retention of the
parent's ability at a fixed budget under a fixed canvas, and none of these
mechanisms was tested here.

\section{Sampling dispersion: per-cell tables}\label{app:disp}

Every score in this paper other than the greedy rows is an estimate from a
finite sample, and the committed values are single draws. We measured the
dispersion directly: three decode seeds
(1234/2345/3456) on each of five committed cells, with the training and the
checkpoints untouched. Greedy generations were token-identical across the three tested decoding
seeds on all 164 problems in each checked cell (score range
\ebOneGreedyRange{}), so the sampler is the only source of movement in these
cells.

Forming the pair ratio within seed and within box gives
\ebTwoRatioBand$\times$. The width comes almost entirely from the
denominator: the in-place arm's three-seed range is \ebTwoDFiveRange{} on a
mean of \ebTwoDFiveMean{}, while the frozen-tower arm contributes
\ebTwoQttRange{} on a mean of \ebTwoQttMean{}. A ratio taken against a
near-floor baseline is not a stable point estimate, and precision on the
numerator does not repair it. The committed scores sit below every per-seed
ratio we measured, so the committed ratio is below the three re-decoded
ratios. Dispersion is
a property of the arm rather than of the budget: at 500\,M the two training
seeds give ranges of \ebOneQttFiveHRange{} and \ebTwoSeedTwoRange{} on the
same task under the same protocol, and it cannot be assumed from one
checkpoint.

\paragraph{Bootstrap intervals over problems.} Decode dispersion measures
one source of uncertainty, the sampler, on a fixed set of problems. The set
itself is a second source, because \heval{} has only 164 problems in it.
Resampling problems with replacement ($10{,}000$ draws) and recomputing
\passat{10} from each cell's per-problem counts gives the 95\,\% intervals
in Table~\ref{tab:boot}. Their half-width is about six points for every
cell above 20, twice the three-point single-draw rule. The two quantities
answer different questions. The decode resampling describes how far a score
moves when the sampler is redrawn on the same problems. The bootstrap
interval bounds how far a score on these 164 problems can sit from the score
on the population they sample. The comparisons this paper draws are paired
on the same problems, so the interval that matters for each is the paired
bootstrap of the difference in the lower block. The pair gap and the dense
gap exclude zero by a wide margin. The $S{=}16$ stage and the second
training seed do not, which is the result reported for both in
\S\ref{sec:dispersion}. Intervals are computed from the committed
per-problem files, and the committed scores are unchanged.

\begin{table}[H]
\caption{\textbf{Bootstrap 95\,\% intervals over the 164 \heval{} problems.}
Upper block: single cells with per-problem counts. Lower
block: paired differences, resampling problems once for both arms; where a
cell has three decode seeds, the per-problem estimate averages them. The
in-place 1\,B row uses the three re-decoded seeds of Table~\ref{tab:disp},
since the committed draw has no per-problem file.}
\label{tab:boot}
\centering\small
\begin{tabular}{lcc}
\toprule
cell & \passat{10} & 95\,\% interval \\
\midrule
frozen tower, 1\,B (\qtt) & 71.60 & [65.6, 77.6] \\
in-place, 1\,B (\dfive), three-seed re-decode & 4.61 & [2.1, 7.7] \\
dense frozen tower, 500\,M (\dps) & 53.13 & [46.5, 59.9] \\
dense in-place, 500\,M (\dpi) & 6.55 & [3.3, 10.2] \\
placement arm, 250\,M & 38.29 & [31.8, 44.7] \\
1.7\,B donor @ 1.136 tok/param & 11.12 & [6.9, 15.8] \\
1.7\,B donor @ 0.568 tok/param & 6.05 & [2.7, 9.9] \\
1.7\,B donor @ 500\,M & 0.61 & [0.0, 1.5] \\
\midrule
comparison & difference & 95\,\% interval \\
\midrule
frozen vs in-place, 1\,B (3 seeds each) & $+62.9$ & [$+56.8$, $+68.8$] \\
dense frozen vs dense in-place, 500\,M & $+46.6$ & [$+40.1$, $+53.2$] \\
$S{=}16$ stage vs $S{=}32$ arm (3 seeds each, Hopper reads) & $+1.9$ & [$-1.9$, $+5.8$] \\
training seed 2 vs seed 1, 500\,M (3 seeds each) & $+1.8$ & [$-2.3$, $+5.8$] \\
1.7\,B donor, 1.136 vs 0.568 tok/param & $+5.1$ & [$+2.1$, $+8.6$] \\
\bottomrule
\end{tabular}
\end{table}

\paragraph{\hevalplus{} and \passat{1} for the pair.} The same committed
generations score on \hevalplus{} \citep{liu2023evalplus}, whose test suites
are stricter, and at \passat{1} (Table~\ref{tab:pairplus}). The pair scores
\qttHEplusTen{} and \dfiveHEplusTen{} \passat{10} on \hevalplus{}; at
\passat{1} the ratio is about \pairPassOneRatio$\times$ against
\pairPassTenRatio$\times$ at \passat{10}, so the advantage is present per
sample and larger there.

\begin{table}[H]
\caption{\textbf{\hevalplus{} and \passat{1} for the pair}, from the committed
generations. The in-place committed cell has no per-problem file, so its
interval is not computed; its three decode seeds give \passat{1} of
1.22--1.52 with intervals inside [0.3, 2.7].}
\label{tab:pairplus}
\centering\footnotesize
\resizebox{\linewidth}{!}{%
\begin{tabular}{lcccc}
\toprule
arm & \passat{10} & \hevalplus{} \passat{10} & \passat{1} [95\,\%] & \hevalplus{} \passat{1} \\
\midrule
\qtt{} (frozen tower) & 71.60 & \qttHEplusTen{} & \qttHEone{} [28.7, 38.5] & 28.38 \\
\dfive{} (in-place) & 6.19 & \dfiveHEplusTen{} & \dfiveHEone{} & 1.49 \\
\bottomrule
\end{tabular}}
\end{table}

\paragraph{GSM8K and MBPP intervals.} The two large tasks were single draws.
Re-scoring the committed generations through the frozen harness and
resampling problems ($10{,}000$ draws) gives Table~\ref{tab:bootlarge}: the
half-width is about two points on GSM8K (1{,}319 problems) and four on MBPP
(500), and the paired frozen-versus-parent gap on GSM8K excludes zero. Every
cell first had to reproduce its committed accuracy to the digit; two MBPP
cells re-scored one problem differently on this host, in opposite directions,
and are shown at their committed values without an interval. The in-place
1\,B arm's generations were not archived, so its committed single-draw values
stand without an interval. The committed scores are unchanged.

\begin{table}[H]
\caption{\textbf{Bootstrap 95\,\% intervals over problems for GSM8K and MBPP},
in percentage points, from the committed generations. $^{\dagger}$Re-scored one
problem in 500 differently from the committed value; no interval.
$^{\ddagger}$The parent's committed figures differ between sources: the parity
files carry 88.02 / 74.40, the results documents the main text quotes carry
\parentGSM{} / \parentMBPP{}, and the re-scored generations give 88.02 /
74.60; every difference lies inside the interval shown. $^{\S}$No problem
was solved in these cells, so resampling problems gives a degenerate
interval. The exact binomial 95\,\% interval for 0 of 500 is [0.0, 0.7].}
\label{tab:bootlarge}
\centering\small
\begin{tabular}{lcc}
\toprule
cell & GSM8K [95\,\%] & MBPP [95\,\%] \\
\midrule
frozen tower, 1\,B (\qtt) & 82.56 [80.5, 84.5] & 53.40 [49.0, 57.8] \\
AR parent (\parent)$^{\ddagger}$ & 88.02 [86.3, 89.8] & 74.40$^{\dagger}$ \\
dense frozen tower, 500\,M (\dps) & 69.90 [67.5, 72.3] & 41.00 [36.8, 45.2] \\
dense in-place, 500\,M (\dpi) & 5.38 [4.2, 6.7] & 0.00 [0.0, 0.0]$^{\S}$ \\
placement arm, 250\,M & 64.29 [61.7, 66.9] & 14.00 [11.0, 17.2] \\
1.7\,B donor @ 1.136 tok/param & 0.83 [0.4, 1.4] & 0.60 [0.0, 1.4] \\
1.7\,B donor @ 500\,M & 2.35 [1.6, 3.2] & 0.00 [0.0, 0.0]$^{\S}$ \\
RND1 at $T{=}0.01$ & 21.00 [18.8, 23.3] & 19.00$^{\dagger}$ \\
\midrule
comparison & difference & 95\,\% interval \\
\midrule
frozen 1\,B vs parent, GSM8K & $-5.5$ & [$-7.6$, $-3.3$] \\
dense in-place vs dense frozen, GSM8K & $-64.5$ & [$-67.2$, $-61.8$] \\
dense in-place vs dense frozen, MBPP & $-41.0$ & [$-45.2$, $-36.8$] \\
\bottomrule
\end{tabular}
\end{table}

Two different Hopper hosts score the same checkpoint's greedy at
\ebTwoCrossBoxGreedy{}, the same to the digit, and a Blackwell host reads
\ebTwoArchShift{} points away on that same cell. So the difference goes with
the architecture, and two hosts of the same architecture agree. Every
three-seed comparison in this appendix is formed within one accelerator
architecture. A shift
measured on one checkpoint says nothing about comparing two different
checkpoints across hosts.

Three decode seeds (1234/2345/3456) per cell, training and checkpoints
untouched, ranges rather than parametric intervals, every seed reported in
Table~\ref{tab:disp}, including anomalous ones. The frozen-tower 1\,B and in-place 1\,B cells were
measured on one Blackwell host so that the per-seed ratio is formed within
box. The 500\,M and $S{=}16$ cells were measured on Hopper hosts. Their
comparators are Hopper reads too. The $S{=}16$ stage is compared with the
$S{=}32$ arm's three Hopper draws, mean \ebOneQttMean{}. The two training
seeds at 500\,M were read on the two Hopper hosts that agree to the digit on
greedy. None of the re-evaluations in
this appendix forms a ratio or difference across the two architectures. The
committed headline and attach-density comparisons do cross them and are read
with that limitation (Appendix~\ref{app:repro}).

\begin{table}[H]
\caption{\textbf{Decode dispersion.} ``Committed'' is the single draw
reported in the main text and is not revised. Greedy generations are token-identical across seeds in every cell
(164/164 problems).}
\label{tab:disp}
\centering\small
\resizebox{\linewidth}{!}{%
\begin{tabular}{llcccc}
\toprule
cell & metric & committed & draws (1234 / 2345 / 3456) & 3-seed mean & range \\
\midrule
frozen-tower 1\,B, $S{=}32$ (Blackwell read) & \heval{} \passat{10} & \qttHEten & 68.36 / 66.60 / 67.46 & \ebTwoQttMean & \ebTwoQttRange \\
frozen-tower 1\,B, $S{=}32$ (Hopper read) & \heval{} \passat{10} & \qttHEten & \ebOneQttSeeds & \ebOneQttMean & \ebOneQttRange \\
in-place 1\,B & \heval{} \passat{10} & \dfiveHEten & 3.66 / 4.64 / 5.52 & \ebTwoDFiveMean & \ebTwoDFiveRange \\
frozen-tower 500\,M, seed 1 & \heval{} \passat{10} & \qttHEtenB & 64.15 / 63.57 / 63.87 & \ebTwoSeedOneMean & \ebOneQttFiveHRange \\
frozen-tower 500\,M, seed 2 & \heval{} \passat{10} & \qttSTwoHEtenB & 65.84 / 64.38 / 66.62 & \ebTwoSeedTwoMean & \ebTwoSeedTwoRange \\
frozen-tower $S{=}16$ & \heval{} \passat{10} & \ssixteenHEten & 70.50 / 69.64 / 71.06 & \ebTwoSsixteenMean & \ebTwoSsixteenRange \\
\midrule
any cell & \heval{} greedy & --- & identical & --- & \ebOneGreedyRange \\
\bottomrule
\end{tabular}}
\end{table}

These are the two comparisons behind the rule in
\S\ref{sec:dispersion}, and this is how far they move. The training-seed
difference at 500\,M goes from $+0.41$ on single draws to
$-\ebTwoSeedDelta$ on three-draw means. The $S{=}32$ against $S{=}16$
difference goes from $+1.94$ to $-\ebTwoStageDelta$. Both arms of each
comparison were read on the same accelerator architecture, and neither
committed number changes.

Deterministic tasks (GSM8K, \mmlupro{}, MBPP at $n{=}1$, temperature 0)
never consult the sampler's generator. Re-running one of them under two
seeds reproduced the committed output byte for byte, so their dispersion is
zero by construction.

The committed baselines are unseeded single draws: the evaluation entry
point called no seeding routine, and where a seed did reach a generator its
base was the interpreter's per-process string hash. This does not invalidate
any committed number, since \passat{10} from $n{=}20$ is a valid estimate
whether or not the draw was seeded, but the exact generations cannot be
regenerated. The hash seed is pinned in every cell reported here and in all
later work.

\section{Interface ablations: detail}\label{app:interface}

\paragraph{Denoiser size, and the quantity that orders it.} We then hold the
tower and the attach fixed and shrink the denoiser, replacing the full
parent copy (0.919\,B trainable) with dense same-family donors. A 0.6\,B
donor (\armbTrainable\,B trainable) at 500\,M tokens reaches \armbHEten{} on
\heval{} \passat{10} and \armbMMLUP{} on \mmlupro{}, having sat at chance
(\armbMMLUPA) at 250\,M. A 1.7\,B donor (\armbTwoTrainable\,B trainable) at
the same 500\,M budget reaches \armbTwoHEten{} and \armbTwoMMLUP{}, and is
\emph{flat at chance} across its entire second 250\,M
(\armbTwoMMLUPA{} to \armbTwoMMLUP{}) while the smaller donor nearly tripled
over the same interval. It reaches this with three times the trainable
parameters and a lower final training loss (\armbTwoLoss{}). Its commit
entropy at 500\,M, \armbTwoEntropy{}, sits within 0.006 of the smaller
donor's \emph{250\,M} value (\armbEntropyA{}), where that donor had already
fallen to \armbEntropyB{} by 500\,M.

Ordered by \emph{tokens per trainable parameter}, the 1.7\,B donor sits at
\tokparamArmbTwoA{} and \tokparamArmbTwoB{}, and the 0.6\,B donor at
\tokparamArmbA{} and \tokparamArmbB{}. Along that order \mmlupro{} reads
\armbTwoMMLUPA{}, \armbTwoMMLUP{}, \armbMMLUPA{} and \armbMMLUP{}.
\heval{} reads \armbTwoHEten{}, \armbHEtenA{} and \armbHEten{} at three of
those points, because the 1.7\,B donor's 250\,M checkpoint was never scored
on it. Only the last point separates from chance on \mmlupro{}, and it is the
only \heval{} score above one point. At its full budget the larger donor has seen fewer
tokens per trainable parameter than the smaller one had at half of its.

At a fixed token budget, the smaller donor scores higher in this
comparison. That comparison mixes donor size with tokens per
trainable parameter. Matching tokens per trainable parameter instead
removes the mixing, and it measures nothing about convergence. Neither
comparison sets a lower bound on the capacity a denoiser needs.

\paragraph{The 1.7\,B donor at matched tokens per trainable parameter.}
Running the 1.7\,B donor to \tokparamArmbB{} tokens per trainable parameter,
the 0.6\,B donor's 500\,M position, holds tokens per trainable parameter
fixed across the two donors. The
run was set up with two co-primary metrics fixed in advance and a conflict rule: if
the two land in bands that point in opposite directions, both are reported
and no size conclusion is drawn. They do. At the matched point the larger
donor reaches \armbXHEten{} on \heval{} \passat{10} against \armbHEten{}
for the smaller donor, inside its \textsc{size-helps} band, and
\armbXMMLUP{} on \mmlupro{} against \armbMMLUP{}, inside its
\textsc{knowledge-flat} band. We draw no conclusion about donor size from
this pair, and we report what each metric shows.

The knowledge trajectory adds four points. In the extended run, \mmlupro{}
was read at four checkpoints, \armbXMMLUPone{}, \armbXMMLUPT{},
\armbXMMLUPthree{} and \armbXMMLUP{} at 0.355, \tokparamArmbA{}, 0.852 and
\tokparamArmbB{} tokens per trainable parameter, with chance at 10 and one
standard error near 1.5. The first of these is a different run from the
donor's own 500\,M run in Table~\ref{tab:donorsize}, which reads
\armbTwoMMLUP{} at the same tokens per parameter: the two runs share the
recipe and differ in learning-rate path (Appendix~\ref{app:prereg}), and both
values sit near chance. The larger donor stays within the neighbourhood of
chance across a threefold increase in adaptation, while the smaller donor
rose from \armbMMLUPA{} to \armbMMLUP{} over the same interval. Additional
training improves the larger donor's \heval{} score, while its \mmlupro{}
score remains below the smaller donor's.

Code generation did switch on. The \heval{} margin over the smaller donor is
\armbXHEtenT{} against \armbHEtenA{} at \tokparamArmbA{} and \armbXHEten{}
against \armbHEten{} at \tokparamArmbB{}, an offset of about five and a half
points that is present at the first matched point and does not grow. The
secondary metrics divide the same way. JSON validity rises from
\armbJSONvalid\% to \armbXJSONvalid\%, CoNLL-JSON micro-F1 from
\armbConllF{} to \armbXConllF{}, and SQuAD-short EM from \armbSquad{} to
\armbXSquad{}; GSM8K (\armbXGSM{}) and MBPP (\armbXMBPP{}) are at floor for
both donors and resolve nothing. At matched adaptation the larger donor
scores higher on the formatting metrics and lower on \mmlupro{}. The
comparison does not isolate donor size, because the learning-rate
trajectories and the hosts differ.

Three caveats attach. The arm's learning-rate path was not schedule-matched to
the smaller donor's. An implementation error ran its first 12{,}140 steps
as a complete cosine to the floor and then re-warmed, so the endpoint
learning rate is the designed one and the path is not
(Appendix~\ref{app:prereg}). The direction of the deviation's effect was not measured. It is a further
limitation of this arm's comparison. And this is one donor size at
one matched point, so it does not locate a capacity floor. The arm also ran
on a different accelerator architecture from the smaller donor, with the
micro-batch halved and gradient accumulation doubled
(Appendix~\ref{app:prereg}).

\begin{table}[H]
\caption{\textbf{Denoiser size at a fixed tower and attach}, ordered by
tokens per trainable parameter (tok/param). The first four rows are the two
donors' own runs; the 800\,M and 1.601\,B rows are the extended run of the
1.7\,B donor, whose own checkpoint at 0.355 tok/param (step 5,059) reads
\armbXMMLUPone{} on \mmlupro{} against \armbTwoMMLUP{} for the separate
500\,M run, the two runs differing in learning-rate path
(Appendix~\ref{app:prereg}).}
\label{tab:donorsize}
\centering\small
\begin{tabular}{lccccc}
\toprule
denoiser & \heval{} \passat{10} & \mmlupro{} & trainable params & tokens & tok/param \\
\midrule
0.6\,B donor @ 250\,M & \armbHEtenA & \armbMMLUPA & \armbTrainable\,B & 250\,M & \tokparamArmbA \\
0.6\,B donor @ 500\,M & \armbHEten & \armbMMLUP & \armbTrainable\,B & 500\,M & \tokparamArmbB \\
1.7\,B donor @ 250\,M & --- & \armbTwoMMLUPA & \armbTwoTrainable\,B & 250\,M & \tokparamArmbTwoA \\
1.7\,B donor @ 500\,M & \armbTwoHEten & \armbTwoMMLUP & \armbTwoTrainable\,B & 500\,M & \tokparamArmbTwoB \\

1.7\,B donor @ 800\,M & \armbXHEtenT & \armbXMMLUPT & \armbTwoTrainable\,B & 800\,M & \tokparamArmbA \\
1.7\,B donor @ 1.601\,B & \armbXHEten & \armbXMMLUP & \armbTwoTrainable\,B & 1.601\,B & \tokparamArmbB \\
\midrule
full parent copy @ 500\,M & \qttHEtenB & 54.2 & 0.919\,B & 500\,M & 0.544 \\
\bottomrule
\end{tabular}
\end{table}

\paragraph{Attach sets.} Density arms use a uniform stride and are strictly
nested: $\{3,7,\dots,47\}\subset\{1,3,\dots,47\}\subset
\{\ell:\ell\not\equiv0\bmod4\}\subset\{0,\dots,47\}$, and every set contains
layer 47. Non-attached layers are canvas-only, with zero-length tower keys
and values. Nesting is what makes the sequence a monotone inclusion chain.

\paragraph{Placement at matched density.} The density arms above change how many
tower layers the denoiser reads, and all of them spread those layers evenly
through the tower. They therefore say nothing about which layers are read.
A fifth arm keeps the density at half and reads the top 24 layers as one
block, $\{24,\dots,47\}$.
Budget, corpus, seed, recipe and harness are identical to the strided arm, and
both ran on the same accelerator architecture.

On the metric the density curve is read on, the two placements differ by
1.8 points: the contiguous arm scores \qnThreeHEten{} on \heval{}
\passat{10} against \qnFiftyHEten{} for the strided arm, and this
single-seed comparison does not establish equivalence. On two of the five
tasks it matters a great deal. \mmlupro{} falls from \qnFiftyMMLUP{} to
\qnThreeMMLUP{}, which is below the quarter-density arm's
\qnTwentyfiveMMLUP{}, and MBPP falls from \qnFiftyMBPP{} to \qnThreeMBPP{},
closer to the quarter-density arm's \qnTwentyfiveMBPP{} than to its own
density. \heval{}+ and GSM8K move by less
(\qnFiftyHEplusten{} to \qnThreeHEplusten{}, and \qnFiftyGSM{} to \qnThreeGSM{}).

The two placements differ by 1.8 \heval{} \passat{10}
points, and this single-seed comparison does not establish equivalence. On
\mmlupro{} and MBPP the contiguous placement scores \qnThreeMMLUP{} and
\qnThreeMBPP{} against \qnFiftyMMLUP{} and \qnFiftyMBPP{}, so density alone
does not fix knowledge decoding, and a uniform stride is doing work that a
contiguous half does not.
And the metric-specificity reported for the density curve holds for placement
too: the density result is a statement about a uniformly strided half
measured on \heval{} \passat{10}, with the uncertainty of the contrasts
unmeasured, and not a general one.

The step-0 routing divergence anticipates this where the gate metric does not.
The contiguous arm reads \qnThreeKL{} against the strided arm's 0.3205, higher
than every density point including zero attachment. That ordering matches what
\mmlupro{} and MBPP show and contradicts what \heval{} \passat{10} shows, which
places a second limit on the surrogate: it is not capability-ordered at low
density, and it is not gate-metric-ordered across placement.

\paragraph{Rounds per block fall monotonically with density.}
Rounds per block fall 13.7, 11.54, 10.13, 8.4 across a quarter, half, three
quarters and full density on \heval{}. The first-round commit fraction reads
3.1--7.3\%, 12.2\%, 14.7\% and 9--22\% at the same four densities. The first
and last are ranges across tasks and they overlap the middle values, so this
metric does not order the arms. Unlike the headline metric, rounds per block
do not plateau between half and three-quarter density: more attached layers
give earlier commits and fewer rounds.

\paragraph{A cheap surrogate that only partly works.} Routing divergence
against the frozen tower at step 0, before any training, is monotone in
density from a quarter upward: 0.476, 0.321, 0.227, 0.130. It is not
capability-ordered at the bottom, since zero attachment reads below a quarter
while scoring far worse when trained, and its shape disagrees with the
trained curve, which steps at half density where the surrogate is close to
linear. It is usable as a coarse ranking signal at or above a quarter density
and is not a substitute for a trained run. One recorded discrepancy stands
unresolved: the quarter-density level did not reproduce an earlier
measurement of the same quantity on the same geometry.

\paragraph{Training loss does not see this damage.} Final training loss
varies by about 0.3 across arms whose \heval{} \passat{10} spans
\qnTwentyfiveHEten{} to \qnDenseHEten{}. The same observation holds across
every arm in this paper and is the reason the gates are defined on held-out
task metrics.

\paragraph{Denoiser-size arms, recorded non-identities.} The 0.6\,B and 1.7\,B donor arms ran on different hosts and different
architectures, a limitation of every comparison between them.
The 1.7\,B arm used a smaller micro-batch with proportionally more gradient
accumulation to fit its host, holding sequences per optimiser step fixed, so
the two arms are not bit-identical in their gradient reduction. Generations
from the 1.7\,B arm are not degenerate: 3,280 completions, mean 560
characters, one empty, fluent and syntactically valid Python that drifts off
task rather than failing to produce text.

\section{Routing under conversion}\label{app:routing}

Expert routing never collapsed in any of our conversions. We logged it in
\routerContexts{} contexts: full fine-tuning, frozen-with-alignment,
block-structured, uniform-state, and the frozen-tower denoiser copy. The
router was trainable in all five. The dead-expert fraction was
\routerDeadFrac{} at every logged step, and utilisation entropy stayed at or
above \routerUtilEntropy{}, over roughly \routerCumTokens\,B supervised
tokens in total.

Swapping the attention mask moves routing before any training happens. At
identical weights at step 0, the divergence is \klSwapFull{} under a
full-bidirectional layout and \klSwapBlock{} under a block-causal layout on
the small parent, and \klSwapQwen{} on the 30B parent. All three reproduce
across hosts to four decimal places. Block structure therefore keeps routing
close to where the AR model had it, and full bidirectionality moves it about
twice as far. We are not aware of prior published measurements of routing
behaviour under \atod{} conversion.

\section{Interventions that did not improve decoding}\label{app:negatives}

Three interventions aimed at faster or better decoding were measured, and
none of them moved the quantity it targeted.

\paragraph{Few-step distillation does not buy sampler rounds at this scale.}
Self-distillation through time was run at two grid depths against the
deployed adaptive sampler, whose operating point is \ltwoRefRounds{} rounds
per block. A grid roughly six times above the operating point gave
\ltwoStageOneRounds{} rounds, and a rung placed directly \emph{on} the
operating point gave \ltwoBRounds{}. The speed metric is insensitive to grid
depth, and both rungs drift marginally upward. Quality was
never the obstacle: the on-operating-point rung scored \ltwoBHEten{} on
\heval{} \passat{10} against a prespecified floor of 62. We ran the second rung to test an
explanation of the first. Uniform ancestral rungs optimise a schedule the
adaptive sampler never visits, so a rung placed on the operating point
should have moved the round count. It did not, so that explanation does not
hold. We report round counts only, because the wall-clock figures from these
runs came from a different serving stack than \S\ref{sec:throughput} and are
not comparable to it.

\paragraph{Self-correction is real and buys nothing.} The uniform-state arm is the only
one of ours with a live self-conditioning module. Switching it off changes
the decode a lot. Revisions per completion fall from 9.7 to 0.6, and final
canvas entropy falls from 0.524 to 0.257. The module is doing exactly what it is
for. The arm still scores 0.0 on code generation. A self-correction channel
is therefore not sufficient for the commit behaviour that makes a diffusion
decoder fast.

\paragraph{More sampler steps buy confidence without quality.} Sweeping the step
cap on that arm at a fixed entropy floor moves final canvas entropy
monotonically, 0.635 at a cap of 16 to 0.363 at 256, at a fivefold cost in
throughput, 101.8 to 20.4 tokens per second. \heval{} \passat{1} and
\passat{10} are 0.0 at every cap. The decoder becomes more confident about
text that does not solve the task.

\section{Prespecified bands and deviations}\label{app:prereg}

Each arm was launched against a document that fixed its numerical bands, the
reading for each band, the launch bars and the stopping rule. The document
was committed before the first training step. We never revise a committed
number. A later result changes the wording and leaves the number as
measured. Four runs departed from what their document specified. We list all four
below.

\begin{enumerate}
\item \emph{Micro-batch on the 1.7\,B donor.} The recipe did not fit the
  memory of its host, so we halved the micro-batch and doubled gradient
  accumulation, holding sequences per optimiser step fixed. Schedule,
  optimiser, freeze set, attachment, data order and seed are untouched. The
  accumulation order differs from the 0.6\,B arm's, so the two arms do not
  reduce gradients bit-identically.
\item \emph{Learning-rate path on the 1.7\,B donor at matched adaptation.}
  We ran it in two parts so that a report point could be read before the last
  15\,h were spent. The scheduler received the first part's step count,
  annealed to the floor over 12{,}140 steps and re-warmed on resumption. The run ended at the designed learning rate and reached it by
  a different path. We did not measure which way this moved the arm
  (Appendix~\ref{app:interface}). We did not rerun the arm, because no
  conclusion in the paper depends on it.
\item \emph{The witness reference and criterion.} The band for the untrained
  frozen-tower sampler was written against the parent's served greedy score,
  53.66. On the frozen harness's own decode path the parent scores 53.05,
  measured and committed before the band was set, and re-decoding it on the
  witness's box gives 52.44. The witness landed on 53.05. It matches the
  committed score, sits 0.61 above the same-box one, and falls outside the
  band, so the run stopped there. Before any later cell ran, we replaced the
  criterion with forced-token agreement against the parent's teacher-forced
  argmax along the witness's own completions, 99.61\% for the witness
  against 99.50\% for the parent's own free-running decode, together with a
  score within 0.5 of the own-decode reference (Appendix~\ref{app:mech}).

  The criterion also covered sampling. The witness's \passat{10} was
  registered to fall inside the parent's paired-bootstrap interval around
  89.23. The parent's cell has
  no per-problem file, so that interval could not be formed and the
  registered check could not be evaluated. The gate document, written after
  the cell was measured, reports instead whether the parent's value lies
  inside the witness's own problem-resample interval. We did not register
  that substitute in advance.
\item \emph{An added baseline row.} The parent initialisation scored under
  each arm's objective (Table~\ref{tab:elbo}) was added after the held-out
  readings, so that each arm's reduction from its own initialisation could
  be reported. It carries no band and changes no reading.
\end{enumerate}

\paragraph{Bands referenced from the text.} The trainable-tower control
(\S\ref{sec:theory}) was read on two metrics. \heval{} \passat{10} at
500\,M was read against 64.04, \emph{helps} at or above 67.0,
\emph{matched} between 61.0 and 67.0, and \emph{hurts} below 61.0.
\mmlupro{} was read against 54.2, \emph{up} at or above 57.2, \emph{flat}
between 51.2 and 57.2, and \emph{down} below 51.2. If the two point in
opposite directions, both are reported and no single conclusion about
freezing is drawn. The witness (Appendix~\ref{app:witness}) was read on
greedy \passat{1} within 0.5 of 53.66, and on sampled \passat{10} inside the
parent's interval around 89.23, which could not be evaluated (item 3
above). No band was placed on the $K$-sweep, and the 250\,M trajectory
points carry no band.

\section{The frozen harness}\label{app:harness}

Every model in this paper, including the parents and the external baseline,
was scored by one harness, fixed before the run and unchanged
throughout. Where a comparison crosses implementations we say so explicitly
and report both sides (\S\ref{sec:protocols}).

\paragraph{Task suites.} We score two suites. The first holds \heval{}
\citep{chen2021humaneval} (164 problems), \heval{}+
\citep{liu2023evalplus}, MBPP \citep{austin2021mbpp} (500), GSM8K
\citep{cobbe2021gsm8k} (1{,}319) and \mmlupro{} \citep{wang2024mmlupro} (500
questions, where chance is 10). The second holds BoolQ
\citep{clark2019boolq}, SQuAD-short exact match \citep{rajpurkar2016squad}
and CoNLL-to-JSON validity \citep{tjong2003conll}. We added the second
suite because its tasks want short answers, and comparing them with the
first suite's longer generations is what makes the output-length effect of
\S\ref{sec:inplace} visible.

The task definitions are fixed in the harness configuration. \heval{}
draws 20 samples per problem at temperature 0.8 and top-$p$ 0.95 with 256
new tokens, plus one greedy sample, and \heval{}+ scores the same
generations. GSM8K is the full test split, greedy, 256 new tokens. MBPP is
the 500 test problems, 3-shot in the format of \citet{austin2021mbpp},
greedy, 256 new tokens. \mmlupro{} is a fixed 500-question subset, drawn as
a seeded uniform sample (seed 42) over the test split in question-id order,
with five shots from the validation split in the question's category,
direct answer, greedy, 32 new tokens. BoolQ is 500 validation questions
drawn with seed 42, 3-shot, 4 new tokens, scored on the first yes/no word,
with unparseable answers counted wrong. SQuAD-short keeps the SQuAD v1.1
validation questions whose reference answer has at most five words, draws
500 with seed 42, and scores exact match under SQuAD normalisation after
3-shot prompting with 16 new tokens. CoNLL-to-JSON draws 500 sentences from
the CoNLL-2003 test split with seed 42; after 3-shot prompting the model
emits one JSON object with PER, ORG and LOC lists in 48 new tokens,
validity is whether the first JSON object parses, and micro-F1 is over
(type, surface) pairs. Dataset revisions are pinned in the configuration.
The questions scored, the generation limits and the scoring rules are also
recorded with the committed per-problem files.

\paragraph{Sampling.} \passat{10} is estimated from $n{=}20$ samples at
temperature 0.8. Greedy \passat{1} is a separate deterministic run and is
reported separately throughout, never mixed into a sampled figure. GSM8K,
\mmlupro{} and MBPP are scored at $n{=}1$ and temperature 0, so the sampler's
generator is never consulted and their dispersion is zero by construction
(Appendix~\ref{app:disp}).

\paragraph{Decode configurations.} The frozen-tower models decode under the
confidence rule at $\gamma=0.8$ with block size $S$ as trained, with the
tower resident. In-place models decode under their native block backend. The
uniform-state arm decodes under its own entropy-bound sampler; the one
cross-application of that sampler to block-trained weights is marked in
Table~\ref{tab:inplace} and is what isolates the decoder from the corruption
process. The external baseline is scored three ways: as published, under our
decoder, and under its own released inference code at a pinned revision, recorded
with the code release.

\paragraph{Parent-agreement diagnostic.} Every cross-protocol table carries
the AR parent measured under the same protocols as the diffusion models. The
parent is the control: a protocol change that moves a diffusion model's score
while leaving the parent's unmoved is evidence of decoder coupling, and does not exclude model-specific
formatting or implementation interactions. In this paper the parent moves by less than three points across the three
protocols, and by at most \parentParityMaxDelta{} points between our decode
and the baseline's own code, on \heval{} \passat{1}. The baseline moves by
far more. That is the
observation \S\ref{sec:protocols} is built on.

\paragraph{Determinism.} The committed scores are single draws from unseeded
runs: the evaluation entry point called no seeding routine, so the exact
generations cannot be regenerated, while \passat{10} from $n{=}20$ remains a
valid estimate. The hash seed is pinned in every cell measured after that
finding, and two byte-identical seeded draws were required before any
dispersion cell ran. Greedy scores
reproduce to the digit across seeds, across processes and, for one
checkpoint, across two physically different hosts of the same architecture.

\section{Reproducibility}\label{app:repro}

\paragraph{One configuration per run, committed, never edited in place.}
Each arm has one configuration file, committed before its launch and never
edited afterwards. A changed recipe becomes a new file, and its header
states what changed against the file it came from. At launch the freeze set
is checked against parameter counts measured from the model itself, and a
mismatch aborts the run.

\paragraph{Corpus.} All 30B-parent arms draw their training data from one
packed corpus of 3\,B supervised tokens, built from FineWeb-Edu (the 10\,BT
sample, 70\,\% by target tokens), MegaMath (megamath-web-pro, 15\,\%) and
the StarCoder slice of OLMoE-mix-0924 (15\,\%, code languages only), at
pinned dataset revisions recorded in the corpus manifest. The matched pair shares one
data order and reads the same prefix of the corpus. The second training seed
uses a different permutation of the same corpus. The parent's pretraining data is not
public, so this is a generic mix and we cannot match the parent's own
distribution.
Documents are tokenized with the parent's tokenizer and packed
document-aware at 256-token alignment, so no canvas block spans a document
boundary. The manifest carries the dataset revisions, per-shard checksums
and the sequence and token counts, 1{,}942{,}309 sequences in all. The two
arms of the matched pair share a bit-identical manifest, which fixes the
data side of that comparison. Every host verifies the manifest before
training starts.

\paragraph{Hosts and environments.} Runs span Hopper and Blackwell hosts with
pinned dependency sets per architecture. Every cross-host comparison in this paper is named as such, and the one
same-checkpoint control that spans two hosts of the same architecture is
reported in Appendix~\ref{app:disp}. The committed headline scores of \S\ref{sec:pair} come from two
architectures. The ratio formed within one host is the three-seed one of
Appendix~\ref{app:disp}, measured on a single Blackwell host. The attach-density
differences of \S\ref{sec:interface} are formed across a Hopper and a
Blackwell host and are reported with that limitation. Every committed cell was scored on the host that trained it, and the
three-seed re-evaluations were scored on the hosts Appendix~\ref{app:disp}
names.

\paragraph{Artefacts.} For the arms marked in Table~\ref{tab:inventory}, the archived record
holds training logs, raw generations and per-problem scores. For the rest it
holds evaluation summaries, calibration records and run headers. The in-place 1\,B arm's committed cells
have summaries only, and its 250\,M--750\,M checkpoints were not archived
(Appendix~\ref{app:mech}). Where generations exist, \passat{k} can be
recomputed at any $k$ and \heval{}+ scored from the same generations without
re-running a model. Checkpoints are mirrored off-host, then downloaded again and checked
against their checksums. Nothing is deleted from a host until that check
passes.

\paragraph{Inventory.} Table~\ref{tab:inventory} lists every trained arm
whose numbers appear in this paper: its training seed,
training host, corruption kernel and canvas block schedule, which artefacts
the archived record holds, and what kind of estimate the reported number is.
The OLMoE arms train on the OLMoE pretraining mixture. Every Qwen-parent
arm trains on the packed corpus described above, the dense pair included.
The Qwen-parent arms train in bf16 with fp32 master weights and optimiser
moments, and the OLMoE full fine-tune used 8-bit paged moments. Every
committed score is a single decode draw. Where Table~\ref{tab:disp} adds
three decode seeds, the three-seed mean is reported beside it, and greedy
scores are exact under the tested conditions.

\begin{table}[H]
\caption{\textbf{Experiment inventory.} The seed is the training
data-order seed. Kernel: compl. = the
independent per-position masking of the in-place arm, blockwise = the
exactly-$k$ masking of the two-tower kernel; the block entry is the canvas schedule. Marks: \checkmark{} in the
archived record; rcpt = off-host copy with a checksum receipt in the record;
n/v = a final checkpoint that exists in the archive but whose off-host
copy was not checksum-verified for this paper; -- = not in the archived record; seeds =
present only for the three decode-seed re-evaluations. Estimate: SD =
committed single decode draw; +3s = three decode seeds also reported
(Table~\ref{tab:disp}). Every committed cell was scored on its training host.}
\label{tab:inventory}
\centering\footnotesize
\resizebox{\linewidth}{!}{%
\begin{tabular}{llllccccl}
\toprule
arm & seed & trained and scored on & kernel, block & log & final & gens & per-prob. & estimate \\
\midrule
OLMoE (i) full fine-tune & 42 & GH200 & uniform mask & -- & n/v & -- & -- & SD \\
OLMoE (ii) frozen + align & 42 & GH200 & uniform mask & -- & n/v & -- & -- & SD \\
OLMoE (iii) block & 42 & GH200 & compl., 32$\to$64 & -- & n/v & -- & -- & SD \\
OLMoE (iv) uniform-state & 42 & GH200 & uniform-state, 64 & -- & n/v & -- & -- & SD \\
\midrule
\dfive{} & 42 & B200 & compl., 32$\to$64 & \checkmark & rcpt & seeds & seeds & SD, +3s \\
\qtt{} & 42 & GH200 & blockwise, 32 & -- & rcpt & \checkmark & \checkmark & SD, +3s \\
\qtt{}, training seed 2 & 2 & B200 & blockwise, 32 & \checkmark & rcpt & \checkmark & \checkmark & SD, +3s \\
$S{=}16$ stage & 42 & GH200 & blockwise, 16 & -- & rcpt & \checkmark & \checkmark & SD, +3s \\
\dpi{} & 42 & GH200 & compl., 32$\to$64 & \checkmark & rcpt & \checkmark & \checkmark & SD \\
\dps{} & 42 & GH200 & blockwise, 32 & \checkmark & rcpt & \checkmark & \checkmark & SD \\
attach 25\,\% & 42 & GH200 & blockwise, 32 & \checkmark & n/v & -- & -- & SD \\
attach 50\,\% & 42 & B200 & blockwise, 32 & \checkmark & n/v & -- & \checkmark & SD \\
attach 75\,\% & 42 & B200 & blockwise, 32 & \checkmark & n/v & -- & \checkmark & SD \\
placement, 50\,\% late & 42 & B200 & blockwise, 32 & \checkmark & rcpt & \checkmark & \checkmark & SD \\
donor 0.6\,B @ 500\,M & 42 & B200 & blockwise, 32 & \checkmark & rcpt & -- & \checkmark & SD \\
donor 1.7\,B @ 500\,M & 42 & GH200 & blockwise, 32 & \checkmark & n/v & \checkmark & \checkmark & SD \\
donor 1.7\,B @ 1.136 tok/param & 42 & GH200 & blockwise, 32 & \checkmark & rcpt & \checkmark & \checkmark & SD \\
distillation rung 32 & 42 & n/r & blockwise, 16 & \checkmark & rcpt & -- & \checkmark & SD \\
distillation rung 8 & 42 & n/r & blockwise, 16 & \checkmark & rcpt & -- & -- & SD \\
trainable tower (\S\ref{sec:theory}) & 42 & H200 & blockwise, 32 & \checkmark & rcpt & \checkmark & \checkmark & SD \\
witness cells (App.~\ref{app:witness}) & untrained init, \qtt{} & GH200 & left-to-right, 32 & \checkmark & n/a & \checkmark & \checkmark & SD \\
\midrule
AR parent (\parent) & -- & -- & -- & -- & -- & \checkmark & -- & SD (parity files) \\
\bottomrule
\end{tabular}}
\end{table}

\begin{table}[H]
\caption{\textbf{Delivered budgets.} Supervised tokens and optimiser steps
as recorded in each run-completion record, which identifies the evaluated
final checkpoint. The frozen-tower 1\,B run's record states its budget as
delivered exactly at 10,113 steps; it shares a bit-identical data manifest
and step count with \dfive{}, so the two arms trained on the same tokens. Nominal budgets are 250\,M, 500\,M and 1\,B; n/r = not in
the record; n/a = the arm did no training. Table~\ref{tab:inventory} gives
what the archived record holds for each arm. All arms were scored with
\texttt{eval\_harness\_v1}. The primary estimate for every reported number is
the committed single draw, by the paper's fixed reporting rule, and the
three-seed mean is its uncertainty companion.}
\label{tab:identifiers}
\centering\footnotesize
\resizebox{\linewidth}{!}{%
\begin{tabular}{ll}
\toprule
arm & delivered budget (tokens / steps) \\
\midrule
\dfive{} & 1,000,075,732 / 10,113 \\
\qtt{} & 1.000\,B / 10,113 \\
\qtt{}, training seed 2 & 499,236,606 / 5,050 \\
$S{=}16$ stage & n/r \\
\dpi{} & 499,313,300 / 5,050 \\
\dps{} & 499,313,300 / 5,050 \\
attach 25\,\% / 50\,\% / 75\,\% & 249,143,808 / 2,520 each \\
placement, 50\,\% late & 249,143,808 / 2,520 \\
donor 0.6\,B / 1.7\,B @ 500\,M & 500,079,173 / 5,059 each \\
donor 1.7\,B @ 1.136 tok/param & 1,600,368,302 / 16,190 \\
distillation rungs 32 / 8 & n/r \\
OLMoE (i)--(iv) & n/r \\
trainable tower & 500,079,173 / 5,059 \\
witness cells & n/a \\
\bottomrule
\end{tabular}}
\end{table}

\paragraph{Cited baseline figures.} The \twotowerBudget{} training budget of
\twotower{} is the figure in that paper's abstract; its training section
reports 1.4\,T tokens across the two training phases alone. The ratios in
\S\ref{sec:pair}, \S\ref{sec:throughput} and \S\ref{sec:related} use the
abstract's figure.

\section{Additional related work}\label{app:related}

\paragraph{Diffusion objectives and samplers.} We build on D3PM's absorbing
and uniform kernels \citep{austin2021d3pm}, the masked-diffusion line of SEDD
and MDLM \citep{lou2024sedd,sahoo2024mdlm}, GIDD and the diffusion duality
for the uniform state \citep{vonrutte2025gidd,sahoo2025duo,deschenaux2026psi},
and BD3-LM's block-mask algebra \citep{arriola2025bd3lm}. Few-step
distillation \citep{deschenaux2025sdtt} and discrete flow matching
\citep{gat2024dfm,shaul2024kinetic} are the standing routes to the throughput
gap of \S\ref{sec:throughput}. Appendix~\ref{app:negatives} measures the
first and finds no gain at this scale.

\end{document}

%% file: axi_macros.tex
\newcommand{\parent}{Qwen3\mbox{-}30B\mbox{-}A3B\mbox{-}Base}
\newcommand{\olmoe}{OLMoE\mbox{-}1B\mbox{-}7B}
\newcommand{\rndone}{RND1}
\newcommand{\qtt}{\textsc{Frozen-Tower-1B}}%
\newcommand{\dfive}{\textsc{InPlace-1B}}%
\newcommand{\dps}{\textsc{Frozen-Tower-Dense}}%
\newcommand{\dpi}{\textsc{InPlace-Dense}}%
\newcommand{\dparent}{Qwen3-4B-Base}%

\newcommand{\twotower}{Nemotron-Labs-TwoTower}

\newcommand{\atod}{A2D}
\newcommand{\passat}[1]{pass@#1}
\newcommand{\heval}{HumanEval}
\newcommand{\mmlupro}{MMLU-Pro}

\newcommand{\hevalplus}{HumanEval+}

%% file: axi_numbers.tex
\newcommand{\parentHEten}{89.23}
\newcommand{\parentHEone}{47.77}
\newcommand{\parentHEgreedy}{53.66}

\newcommand{\parentGSM}{87.19}
\newcommand{\parentMMLUP}{55.40}
\newcommand{\parentMBPP}{74.60}
\newcommand{\parentBoolq}{86.8}
\newcommand{\parentJSONvalid}{99.0}

\newcommand{\olmoeHEten}{22.97}

\newcommand{\ponHEten}{1.01}%
\newcommand{\ptwoHEten}{1.99}%
\newcommand{\pthreeHEten}{1.51}%
\newcommand{\inplaceKnowledgeRet}{77}%

\newcommand{\qttHEten}{71.60}
\newcommand{\qttHEgreedy}{42.07}
\newcommand{\qttHEgreedyA}{22.56}%
\newcommand{\qttHEgreedyB}{44.51}%

\newcommand{\qttGSM}{82.56}
\newcommand{\qttMMLUP}{54.60}
\newcommand{\qttMBPP}{53.40}
\newcommand{\qttJSONvalid}{96.8}

\newcommand{\qttBoolq}{87.2}%
\newcommand{\qttSquad}{60.4}
\newcommand{\qttTokPerFwd}{3.8}
\newcommand{\qttBudget}{1.000\,B}
\newcommand{\qttHEtenA}{47.11}%
\newcommand{\qttHEtenB}{64.04}%
\newcommand{\qttSTwoHEtenB}{63.63}%
\newcommand{\rTwoSpreadMid}{11}%
\newcommand{\qttSTwoMMLUP}{54.2}%
\newcommand{\qttSTwoJSONvalid}{94.2}%

\newcommand{\rndoneHEtenOurs}{18.90}
\newcommand{\rndoneGSMOurs}{21.53}
\newcommand{\rndoneMBPPOurs}{18.20}
\newcommand{\rndoneMMLUPOurs}{43.40}
\newcommand{\rndoneGSMPub}{80.0}
\newcommand{\rndoneMBPPPub}{65.4}

\newcommand{\rndoneBudget}{500\,B}
\newcommand{\rndoneHEtenParity}{90.63}%
\newcommand{\rndoneHEtenParityCI}{$90.6 \pm 0.1$}%
\newcommand{\rndoneHEgreedyParity}{14.63}%
\newcommand{\rndoneHEoneParity}{67.10}
\newcommand{\rndoneGSMParity}{9.93}%
\newcommand{\rndoneMBPPParity}{12.00}%

\newcommand{\dfiveHEten}{6.19}%
\newcommand{\dfiveGSM}{7.96}
\newcommand{\dfiveMMLUP}{45.4}%
\newcommand{\dfiveMBPP}{0.4}
\newcommand{\dfiveSquad}{46.0}%
\newcommand{\dfiveJSONvalid}{31.8}%
\newcommand{\dfiveBoolq}{86.0}%
\newcommand{\parentSquad}{37.6}%

\newcommand{\ssixteenHEten}{69.66}%
\newcommand{\ssixteenBudget}{150\,M}
\newcommand{\qttFirstCommit}{21.6}%
\newcommand{\ssixteenFirstCommit}{42.5}%
\newcommand{\towerOverheadGB}{1.9}%

\newcommand{\latOneLo}{0.134}%
\newcommand{\latOneHi}{0.168}%
\newcommand{\latOneGainLo}{2.09}%
\newcommand{\latOneGainHi}{2.34}%
\newcommand{\latOneStepBefore}{91.30}%
\newcommand{\latOneStepAfter}{16.13}%
\newcommand{\latOneStepGain}{5.66}%
\newcommand{\latOneTrafficQtt}{1.32}%
\newcommand{\latOneTrafficAR}{1.69}%
\newcommand{\latOneTrafficPct}{22}%
\newcommand{\latOneHBMFloor}{2.5}%
\newcommand{\latOneSinkShare}{85.6}%
\newcommand{\latOneStepGainProtLo}{2.97}%
\newcommand{\latOneStepGainProtHi}{3.39}%
\newcommand{\latOneBaseLo}{0.057}%
\newcommand{\latOneBaseHi}{0.080}%

\newcommand{\klSwapFull}{0.211}%
\newcommand{\klSwapBlock}{0.121}%
\newcommand{\klSwapQwen}{0.146}%

\newcommand{\twotowerRetention}{98.7}
\newcommand{\twotowerSpeedup}{2.42}
\newcommand{\twotowerBudget}{2.1\,T}%

\newcommand{\dpsHEten}{53.13}

\newcommand{\dpsHEgreedy}{37.80}

\newcommand{\dpsGSM}{69.90}
\newcommand{\dpsMMLUP}{45.60}
\newcommand{\dpsMBPP}{41.00}
\newcommand{\dpsBoolq}{86.60}
\newcommand{\dpsSquad}{76.00}

\newcommand{\dpsJSONvalid}{96.60}
\newcommand{\dpsRatio}{0.615}
\newcommand{\dpsRetPct}{61.5}%
\newcommand{\dpsBudget}{500\,M}
\newcommand{\dpiHEten}{6.55}
\newcommand{\dpiHEtenA}{6.43}%
\newcommand{\dpiHEgreedy}{0.61}
\newcommand{\dpiGSM}{5.38}
\newcommand{\dpiMMLUP}{32.00}
\newcommand{\dpiMBPP}{0.00}
\newcommand{\dpiBoolq}{80.20}
\newcommand{\dpiSquad}{42.60}
\newcommand{\dpiJSONvalid}{9.80}
\newcommand{\dpiRatio}{0.076}
\newcommand{\dpiRetPct}{7.6}
\newcommand{\dpiCeil}{0.15}%
\newcommand{\dpiMMLUPRet}{68.7}%
\newcommand{\densePairSep}{8.1}%
\newcommand{\dparentHEten}{86.42}

\newcommand{\dparentHEgreedy}{59.76}

\newcommand{\dparentGSM}{81.88}
\newcommand{\dparentMMLUP}{46.60}
\newcommand{\dparentMBPP}{67.60}
\newcommand{\dparentBoolq}{86.80}
\newcommand{\dparentSquad}{81.20}

\newcommand{\dparentJSONvalid}{98.20}

\newcommand{\armbHEten}{5.51}
\newcommand{\armbHEtenA}{0.61}%

\newcommand{\armbMMLUP}{22.0}
\newcommand{\armbMMLUPA}{8.2}%

\newcommand{\armbSquad}{2.2}
\newcommand{\armbConllF}{0.0}
\newcommand{\armbJSONvalid}{15}
\newcommand{\armbTrainable}{0.44}%

\newcommand{\ebOneQttSeeds}{67.51 / 67.90 / 70.07}%
\newcommand{\ebOneQttMean}{68.49}
\newcommand{\ebOneQttRange}{2.56}
\newcommand{\ebOneQttFiveHRange}{0.58}%
\newcommand{\ebOneGreedyRange}{0.00}%
\newcommand{\ebOneRetLo}{76}%
\newcommand{\ebOneRetHi}{80}%

\newcommand{\ebTwoRatioBand}{12.2--18.7}
\newcommand{\ebTwoDFiveMean}{4.61}%
\newcommand{\ebTwoDFiveRange}{1.86}
\newcommand{\ebTwoQttMean}{67.47}%
\newcommand{\ebTwoQttRange}{1.76}
\newcommand{\ebTwoSeedOneMean}{63.86}%
\newcommand{\ebTwoSeedTwoMean}{65.61}%
\newcommand{\ebTwoSeedTwoRange}{2.24}
\newcommand{\ebTwoSeedDelta}{1.75}%
\newcommand{\ebTwoSsixteenMean}{70.40}%
\newcommand{\ebTwoSsixteenRange}{1.42}
\newcommand{\ebTwoStageDelta}{1.91}%
\newcommand{\ebTwoCrossBoxGreedy}{42.68}%
\newcommand{\ebTwoArchShift}{1.83}%

\newcommand{\qnTwentyfiveHEten}{19.60}%
\newcommand{\qnFiftyHEten}{40.09}%
\newcommand{\qnSeventyfiveHEten}{40.94}%
\newcommand{\qnDenseHEten}{47.11}%
\newcommand{\qnFiftyStep}{20.49}%
\newcommand{\qnSeventyfiveStep}{0.85}%
\newcommand{\qnDenseStep}{6.17}%
\newcommand{\qnFiftyMMLUP}{57.00}
\newcommand{\qnSeventyfiveMMLUP}{56.00}
\newcommand{\qnDenseMMLUP}{54.60}
\newcommand{\qnTwentyfiveMMLUP}{43.00}
\newcommand{\qnFiftyGSM}{66.72}
\newcommand{\qnFiftyHEplusten}{35.97}%
\newcommand{\qnSeventyfiveGSM}{73.77}
\newcommand{\qnDenseGSM}{77.33}
\newcommand{\qnTwentyfiveGSM}{50.49}
\newcommand{\qnFiftyMBPP}{35.40}
\newcommand{\qnSeventyfiveMBPP}{39.80}
\newcommand{\qnDenseMBPP}{43.60}
\newcommand{\qnTwentyfiveMBPP}{6.60}
\newcommand{\qnGsmGap}{10.6}%
\newcommand{\qnSeedSpread}{11}%

\newcommand{\armbTwoHEten}{0.61}
\newcommand{\armbTwoMMLUP}{9.2}
\newcommand{\armbTwoMMLUPA}{9.4}%

\newcommand{\armbTwoTrainable}{1.409}%
\newcommand{\armbTwoLoss}{6.58}%
\newcommand{\armbTwoEntropy}{0.8515}%
\newcommand{\armbEntropyA}{0.8571}%
\newcommand{\armbEntropyB}{0.7924}%
\newcommand{\tokparamArmbA}{0.568}%
\newcommand{\tokparamArmbB}{1.136}%
\newcommand{\tokparamArmbTwoA}{0.177}%
\newcommand{\tokparamArmbTwoB}{0.355}%

\newcommand{\rndoneGSMTemp}{21.0}
\newcommand{\rndoneMBPPTemp}{19.0}
\newcommand{\rndoneMMLUPTemp}{41.8}

\newcommand{\parentHEtenParity}{89.20}
\newcommand{\parentHEoneParity}{46.13}
\newcommand{\parentHEgreedyParity}{53.05}
\newcommand{\parentMBPPParity}{74.4}
\newcommand{\parentGSMParity}{88.02}
\newcommand{\parentParityMaxDelta}{1.64}%

\newcommand{\ltwoRefRounds}{4.91}%
\newcommand{\ltwoStageOneRounds}{4.94}%
\newcommand{\ltwoBRounds}{4.99}%
\newcommand{\ltwoBHEten}{71.94}%

\newcommand{\routerContexts}{five}
\newcommand{\routerDeadFrac}{0.0}
\newcommand{\routerUtilEntropy}{0.99}
\newcommand{\routerCumTokens}{2.75}%

\newcommand{\qnThreeHEten}{38.29}
\newcommand{\qnThreeMMLUP}{42.40}
\newcommand{\qnThreeGSM}{64.29}
\newcommand{\qnThreeMBPP}{14.00}

\newcommand{\qnThreeHEplusten}{31.55}
\newcommand{\qnThreeKL}{0.7165}%
\newcommand{\armbXHEten}{11.12}%
\newcommand{\armbXMMLUP}{13.4}%
\newcommand{\armbXHEtenT}{6.05}%
\newcommand{\armbXMMLUPT}{11.6}%
\newcommand{\armbXMMLUPone}{12.0}%
\newcommand{\armbXMMLUPthree}{10.4}%
\newcommand{\armbXGSM}{0.83}%
\newcommand{\armbXMBPP}{0.6}%
\newcommand{\armbXSquad}{7.8}%
\newcommand{\armbXConllF}{7.6}%
\newcommand{\armbXJSONvalid}{81.4}%

\newcommand{\rndoneRetLo}{74}
\newcommand{\rndoneRetHi}{79}

\newcommand{\ttHEten}{63.83}%
\newcommand{\ttMMLUP}{40.8}%
\newcommand{\ttHEtenT}{47.25}%
\newcommand{\ttMMLUPT}{41.2}%
\newcommand{\ttMMLUPTwoFifty}{54.6}
\newcommand{\ttHEplusTen}{57.10}%
\newcommand{\ttGSM}{77.18}%
\newcommand{\ttMBPP}{49.40}%
\newcommand{\ttPeakDelta}{13.8}%
\newcommand{\ttStepFactor}{1.43}%
\newcommand{\ttHEdiff}{-0.21}

\newcommand{\ttMMLUPdiff}{-13.4}
\newcommand{\ttMMLUPdiffLo}{-19.5}
\newcommand{\ttMMLUPdiffHi}{-7.3}
\newcommand{\ttMMLUPdropAbs}{13}
\newcommand{\qttGSMfiveH}{79.76}
\newcommand{\qttMBPPfiveH}{45.8}
\newcommand{\qttHEplusfiveH}{57.66}

\newcommand{\driftKmean}{0.33}
\newcommand{\driftVmean}{0.44}
\newcommand{\driftMax}{0.60}
\newcommand{\driftCosMean}{0.967}
\newcommand{\driftCosMin}{0.946}
\newcommand{\elboInplace}{4.55}
\newcommand{\elboFrozen}{4.68}
\newcommand{\elboInplaceFull}{4.547}
\newcommand{\elboFrozenFull}{4.685}
\newcommand{\elboInplaceSeq}{4.619}
\newcommand{\elboFrozenSeq}{4.758}
\newcommand{\elboGap}{0.138}
\newcommand{\elboPairedMean}{0.139}
\newcommand{\elboPairedSd}{0.581}
\newcommand{\elboInplaceWorseFrac}{35.2}
\newcommand{\elboParentDfive}{16.984}
\newcommand{\elboParentQtt}{16.599}
\newcommand{\elboImprInplace}{12.44}
\newcommand{\elboImprFrozen}{11.91}
\newcommand{\elboImprInplaceSd}{2.34}
\newcommand{\elboImprFrozenSd}{1.99}
\newcommand{\witParity}{42.07}
\newcommand{\witKone}{53.05}%
\newcommand{\witParentSameBox}{52.44}%
\newcommand{\witScoreGap}{0.61}%
\newcommand{\witKtwo}{0.00}
\newcommand{\witTrainedKfour}{0.00}%
\newcommand{\witKoneSampled}{89.56}%

\newcommand{\witKfourSampled}{0.00}
\newcommand{\witAgree}{99.61}%
\newcommand{\witParentAgree}{99.50}%
\newcommand{\witIntLo}{85.21}%
\newcommand{\witIntHi}{93.50}
\newcommand{\witSupOnly}{0.61}%
\newcommand{\witLrOnly}{7.32}%
\newcommand{\witEntKone}{0.33}%
\newcommand{\witEntKmax}{2.42}
\newcommand{\qttHEplusTen}{63.29}
\newcommand{\dfiveHEplusTen}{6.19}
\newcommand{\qttHEone}{33.54}
\newcommand{\dfiveHEone}{1.65}
\newcommand{\pairPassOneRatio}{20}
\newcommand{\pairPassTenRatio}{11.6}

%% file: figures/axi_fig_arch_twotower.tex
\begin{figure}[t]
\centering
\begin{tikzpicture}[
  scale=0.8, transform shape,
  font=\small,
  comp/.style={draw, rounded corners=1pt, minimum width=3.5cm,
    minimum height=0.52cm, inner sep=2pt, align=center},
  frozen/.style={comp, fill=blue!12, draw=blue!50!black},
  train/.style={comp, fill=orange!25, draw=orange!60!black},
  shared/.style={draw=green!55!black, dashed, line width=1pt},
  lbl/.style={font=\footnotesize\itshape},
  flow/.style={-{Stealth[length=2mm]}, thick},
  kv/.style={-{Stealth[length=1.8mm]}, densely dashed, blue!50!black},
  node distance=0.26cm and 3.1cm
]
\node[frozen] (t-att) {causal self-attn};
\node[frozen, above=of t-att] (t-ffn) {MoE experts};
\node[font=\small, above=0.28cm of t-ffn] (t-dots) {$\vdots\;\times L$};
\begin{scope}[on background layer]
\node[draw=blue!50!black, rounded corners=3pt, fill=blue!4, inner sep=7pt,
  fit=(t-att)(t-dots),
  label={[lbl, blue!50!black]above:{\textbf{Context tower} (frozen parent copy)}}]
  (tower) {};
\end{scope}
\node[train, right=of t-att, align=center] (d-att) {bidirectional self-attn\\
  {\scriptsize over $[\,$tower KV $\Vert$ canvas KV$\,]$}};
\node[train, above=of d-att] (d-rt) {router \& norms};
\node[frozen, above=of d-rt] (d-ffn) {MoE experts};
\node[font=\small, above=0.28cm of d-ffn] (d-dots) {$\vdots\;\times L$};
\begin{scope}[on background layer]
\node[draw=orange!60!black, rounded corners=3pt, fill=orange!4, inner sep=7pt,
  fit=(d-att)(d-dots),
  label={[lbl, orange!60!black]above:{\textbf{Denoiser} (trainable copy)}}]
  (deno) {};
\end{scope}
\node[frozen, minimum width=10.6cm,
  below=0.55cm of $(tower.south)!0.5!(deno.south)$]
  (emb) {token embeddings (single copy)};
\draw[flow] (emb.north -| tower) -- (tower.south);
\draw[flow] (emb.north -| deno) -- (deno.south);
\node[shared, rounded corners=2pt, inner sep=2.5pt, fit=(t-ffn)] {};
\node[shared, rounded corners=2pt, inner sep=2.5pt, fit=(d-ffn)] {};
\node[shared, rounded corners=2pt, inner sep=2.5pt, fit=(emb)] {};
\draw[shared, {Stealth[length=1.8mm]}-{Stealth[length=1.8mm]}]
  (t-ffn.east) -- (d-ffn.west)
  node[pos=0.42, above=3pt, font=\scriptsize, green!45!black,
  fill=white, inner sep=1.5pt] {same tensors};
\draw[kv] (t-att.east) -- (d-att.west)
  node[pos=0.45, below=3pt, font=\scriptsize, blue!50!black,
  fill=white, inner sep=1.5pt, align=center]
  {layer-aligned KV\\{\scriptsize concatenated; no new parameters}};
\node[lbl, below=0.35cm of emb, xshift=-2.7cm] (in-clean)
  {committed prefix (clean)};
\node[lbl, below=0.35cm of emb, xshift=2.7cm] (in-canvas)
  {canvas block (masked)};
\draw[flow] (in-clean) -- (in-clean |- emb.south);
\draw[flow] (in-canvas) -- (in-canvas |- emb.south);
\node[lbl, anchor=west] (out)
  at ([xshift=0.35cm, yshift=0.85cm]deno.north east) {denoised block};
\draw[flow, rounded corners=3pt] (deno.east) -| (out.south);
\draw[flow, rounded corners=4pt] (out.west) -| ([xshift=-0.6cm]tower.west)
  |- ([yshift=0.15cm]in-clean.west)
  node[pos=0.25, left, font=\scriptsize, align=center]
  {commit;\\next block};
\matrix[draw=black!30, rounded corners=2pt, inner sep=4pt, font=\scriptsize,
  column sep=3pt, row sep=2pt, below=0.5cm of in-clean, xshift=-0.4cm,
  matrix of nodes, nodes={anchor=west}] {
  |[frozen, minimum width=0.5cm, minimum height=0.28cm]| & frozen \\
  |[train, minimum width=0.5cm, minimum height=0.28cm]| & trainable \\
  |[shared, minimum width=0.5cm, minimum height=0.28cm]| &
    aliased: one copy in memory \\
};
\node[draw=black!30, rounded corners=2pt, inner sep=4pt, font=\scriptsize,
  align=left, below=0.5cm of in-canvas, xshift=0.3cm] {
  weight residency $=$ parent $+\,\towerOverheadGB$\,GB\\
  (98 tensors aliased; 433 tower-own)};
\end{tikzpicture}
\caption{\textbf{The frozen-tower architecture (\qtt).} The frozen context tower is a
bit-exact parent copy running causal attention over the committed prefix; the
denoiser is a second copy of the same architecture, bidirectional over the
masked canvas block, reading the tower through layer-aligned cross-attention \citep{reda2026twotower}, implemented as key/value
concatenation into the denoiser's own attention, so the attach
introduces \emph{no} parameters. One layer is expanded per
tower; computation flows upward. Under the frozen-experts recipe only
attention, router, and norms train (orange). Because
experts and embeddings stay bit-identical to the parent, tower and denoiser
\emph{alias} them (green): the two-tower system holds
parent~$+\,\towerOverheadGB$\,GB of weights at inference, under $2\times$ parent
(\S\ref{sec:throughput}).}
\label{fig:arch}
\end{figure}

%% file: figures/axi_fig_radar.tex
\begin{figure}[t]
\centering
\begin{tikzpicture}
\begin{polaraxis}[
  width=6.2cm,
  xtick={90, 141.43, 192.86, 244.29, 295.71, 347.14, 38.57},
  xticklabels={BoolQ, MMLU-Pro, SQuAD EM, JSON valid, GSM8K, MBPP, HumanEval},
  xticklabel style={font=\footnotesize},
  ymin=0, ymax=100,
  ytick={20,40,60,80,100},
  yticklabel style={font=\tiny, anchor=east},
  grid=both,
  major grid style={black!15},
  legend style={at={(0.5,-0.12)}, anchor=north, legend columns=3,
    font=\footnotesize, draw=black!30},
]
\addplot+[black, dashed, thick, mark=none] coordinates {
  (90,\parentBoolq) (141.43,\parentMMLUP) (192.86,\parentSquad)
  (244.29,\parentJSONvalid) (295.71,\parentGSM) (347.14,\parentMBPP)
  (38.57,\parentHEten) (90,\parentBoolq)};
\addlegendentry{\parent{} (parent)}
\addplot+[teal, thick, mark=*, mark size=1.2pt, fill=teal, fill opacity=0.12]
  coordinates {
  (90,\qttBoolq) (141.43,\qttMMLUP) (192.86,\qttSquad)
  (244.29,\qttJSONvalid) (295.71,\qttGSM) (347.14,\qttMBPP)
  (38.57,\qttHEten) (90,\qttBoolq)};
\addlegendentry{\qtt}
\addplot+[red!70!black, thick, mark=square*, mark size=1.1pt,
  fill=red!70!black, fill opacity=0.12] coordinates {
  (90,\dfiveBoolq) (141.43,\dfiveMMLUP) (192.86,\dfiveSquad)
  (244.29,\dfiveJSONvalid) (295.71,\dfiveGSM) (347.14,\dfiveMBPP)
  (38.57,\dfiveHEten) (90,\dfiveBoolq)};
\addlegendentry{\dfive}
\end{polaraxis}
\end{tikzpicture}
\caption{\textbf{The output-length-dependent degradation, at a glance} (all axes are
percentages on a common $0$--$100$ scale, frozen harness; ordered
counterclockwise by output length).
Both conversions stay close to the parent on the knowledge and
short-output axes (BoolQ, \mmlupro, SQuAD), though \dfive{} is ten
points down on \mmlupro{} and SQuAD EM sits above the parent for the
formatting reason given in \S\ref{sec:pair}. \dfive{} (red) then collapses on the longer-output tasks (JSON validity \dfiveJSONvalid, GSM8K \dfiveGSM, MBPP
\dfiveMBPP{}), while \qtt{} (teal), the frozen-tower configuration, holds at or near the parent on the knowledge and
structured-output axes, at 95\% on GSM8K, and at 72--80\% of it on the
two code axes.}
\label{fig:radar}
\end{figure}

%% file: axi_main.bbl
\begin{thebibliography}{32}
\providecommand{\natexlab}[1]{#1}
\providecommand{\url}[1]{\texttt{#1}}
\expandafter\ifx\csname urlstyle\endcsname\relax
  \providecommand{\doi}[1]{doi: #1}\else
  \providecommand{\doi}{doi: \begingroup \urlstyle{rm}\Url}\fi

\bibitem[Arriola et~al.(2025)Arriola, Gokaslan, Chiu, Yang, Qi, Han, Sahoo, and
  Kuleshov]{arriola2025bd3lm}
Marianne Arriola, Aaron Gokaslan, Justin~T Chiu, Zhihan Yang, Zhixuan Qi, Jiaqi
  Han, Subham~Sekhar Sahoo, and Volodymyr Kuleshov.
\newblock Block diffusion: Interpolating between autoregressive and diffusion
  language models.
\newblock In \emph{The Thirteenth International Conference on Learning
  Representations}, 2025.
\newblock URL \url{https://arxiv.org/abs/2503.09573}.

\bibitem[Austin et~al.(2021{\natexlab{a}})Austin, Johnson, Ho, Tarlow, and
  van~den Berg]{austin2021d3pm}
Jacob Austin, Daniel~D Johnson, Jonathan Ho, Daniel Tarlow, and Rianne van~den
  Berg.
\newblock Structured denoising diffusion models in discrete state-spaces.
\newblock In \emph{NeurIPS}, 2021{\natexlab{a}}.

\bibitem[Austin et~al.(2021{\natexlab{b}})Austin, Odena, Nye, Bosma,
  Michalewski, Dohan, Jiang, Cai, Terry, Le, and Sutton]{austin2021mbpp}
Jacob Austin, Augustus Odena, Maxwell Nye, Maarten Bosma, Henryk Michalewski,
  David Dohan, Ellen Jiang, Carrie Cai, Michael Terry, Quoc Le, and Charles
  Sutton.
\newblock Program synthesis with large language models.
\newblock \emph{arXiv preprint arXiv:2108.07732}, 2021{\natexlab{b}}.

\bibitem[Bie et~al.(2025)Bie, Cao, Chen, Du, Gong, Gong, Gu, Hu, Huang, Lan,
  Li, Li, Li, Li, Liu, Liu, Lu, Lu, Ma, Tan, Wei, Wen, Xing, Zhang, Zhao,
  Zheng, Zhou, Zhou, Zhou, Zhu, and Zhuang]{llada2025v2}
Tiwei Bie, Maosong Cao, Kun Chen, Lun Du, Mingliang Gong, Zhuochen Gong, Yanmei
  Gu, Jiaqi Hu, Zenan Huang, Zhenzhong Lan, Chengxi Li, Chongxuan Li, Jianguo
  Li, Zehuan Li, Huabin Liu, Lin Liu, Guoshan Lu, Xiaocheng Lu, Yuxin Ma,
  Jianfeng Tan, Lanning Wei, Ji-Rong Wen, Yipeng Xing, Xiaolu Zhang, Junbo
  Zhao, Da~Zheng, Jun Zhou, Junlin Zhou, Zhanchao Zhou, Liwang Zhu, and Yihong
  Zhuang.
\newblock {LLaDA2.0}: Scaling up diffusion language models to 100{B}.
\newblock \emph{arXiv preprint arXiv:2512.15745}, 2025.

\bibitem[Chen et~al.(2021)Chen, Tworek, Jun, Yuan, Pinto, Kaplan, Edwards,
  Burda, Joseph, Brockman, et~al.]{chen2021humaneval}
Mark Chen, Jerry Tworek, Heewoo Jun, Qiming Yuan, Henrique Ponde de~Oliveira
  Pinto, Jared Kaplan, Harri Edwards, Yuri Burda, Nicholas Joseph, Greg
  Brockman, et~al.
\newblock Evaluating large language models trained on code.
\newblock \emph{arXiv preprint arXiv:2107.03374}, 2021.

\bibitem[Clark et~al.(2019)Clark, Lee, Chang, Kwiatkowski, Collins, and
  Toutanova]{clark2019boolq}
Christopher Clark, Kenton Lee, Ming-Wei Chang, Tom Kwiatkowski, Michael
  Collins, and Kristina Toutanova.
\newblock {BoolQ}: Exploring the surprising difficulty of natural yes/no
  questions.
\newblock In \emph{Proceedings of NAACL-HLT 2019}, 2019.

\bibitem[Cobbe et~al.(2021)Cobbe, Kosaraju, Bavarian, Chen, Jun, Kaiser,
  Plappert, Tworek, Hilton, Nakano, Hesse, and Schulman]{cobbe2021gsm8k}
Karl Cobbe, Vineet Kosaraju, Mohammad Bavarian, Mark Chen, Heewoo Jun, Lukasz
  Kaiser, Matthias Plappert, Jerry Tworek, Jacob Hilton, Reiichiro Nakano,
  Christopher Hesse, and John Schulman.
\newblock Training verifiers to solve math word problems.
\newblock \emph{arXiv preprint arXiv:2110.14168}, 2021.

\bibitem[Deschenaux \& Gulcehre(2025)Deschenaux and
  Gulcehre]{deschenaux2025sdtt}
Justin Deschenaux and Caglar Gulcehre.
\newblock Beyond autoregression: Fast {LLM}s via self-distillation through
  time.
\newblock In \emph{ICLR}, 2025.
\newblock arXiv:2410.21035.

\bibitem[Deschenaux et~al.(2026)Deschenaux, Gulcehre, and
  Sahoo]{deschenaux2026psi}
Justin Deschenaux, Caglar Gulcehre, and Subham Sahoo.
\newblock The diffusion duality, chapter {II}: {Psi-Samplers}.
\newblock \emph{arXiv preprint arXiv:2602.21185}, 2026.

\bibitem[Gat et~al.(2024)Gat, Remez, Shaul, Kreuk, Chen, Synnaeve, Adi, and
  Lipman]{gat2024dfm}
Itai Gat, Tal Remez, Neta Shaul, Felix Kreuk, Ricky T.~Q. Chen, Gabriel
  Synnaeve, Yossi Adi, and Yaron Lipman.
\newblock Discrete flow matching.
\newblock In \emph{NeurIPS}, 2024.
\newblock arXiv:2407.15595.

\bibitem[Gong et~al.(2025)Gong, Agarwal, Zhang, Ye, Zheng, Li, An, Zhao, Bi,
  Han, Peng, and Kong]{gong2025diffullama}
Shansan Gong, Shivam Agarwal, Yizhe Zhang, Jiacheng Ye, Lin Zheng, Mukai Li,
  Chenxin An, Peilin Zhao, Wei Bi, Jiawei Han, Hao Peng, and Lingpeng Kong.
\newblock Scaling diffusion language models via adaptation from autoregressive
  models.
\newblock In \emph{The Thirteenth International Conference on Learning
  Representations}, 2025.
\newblock URL \url{https://openreview.net/forum?id=j1tSLYKwg8}.

\bibitem[Kumar et~al.(2022)Kumar, Raghunathan, Jones, Ma, and
  Liang]{kumar2022finetuning}
Ananya Kumar, Aditi Raghunathan, Robbie~Matthew Jones, Tengyu Ma, and Percy
  Liang.
\newblock Fine-tuning can distort pretrained features and underperform
  out-of-distribution.
\newblock In \emph{International Conference on Learning Representations}, 2022.
\newblock URL \url{https://openreview.net/forum?id=UYneFzXSJWh}.

\bibitem[Liu et~al.(2023)Liu, Xia, Wang, and Zhang]{liu2023evalplus}
Jiawei Liu, Chunqiu~Steven Xia, Yuyao Wang, and Lingming Zhang.
\newblock Is your code generated by {ChatGPT} really correct? {Rigorous}
  evaluation of large language models for code generation.
\newblock In \emph{Advances in Neural Information Processing Systems 36
  (NeurIPS 2023)}, 2023.

\bibitem[Lou et~al.(2024)Lou, Meng, and Ermon]{lou2024sedd}
Aaron Lou, Chenlin Meng, and Stefano Ermon.
\newblock Discrete diffusion modeling by estimating the ratios of the data
  distribution.
\newblock In \emph{ICML}, 2024.

\bibitem[Ma et~al.(2026)Ma, Xiao, Li, and Zhang]{ma2026fluid}
Xiangyu Ma, Teng Xiao, Zuchao Li, and Lefei Zhang.
\newblock From {AR} to diffusion: Efficiently adapting large language models
  with strictly causal and elastic horizons.
\newblock In \emph{Proceedings of ACL 2026}, 2026.
\newblock arXiv:2605.27387.

\bibitem[Muennighoff et~al.(2025)Muennighoff, Soldaini, Groeneveld, Lo,
  Morrison, Min, Shi, Walsh, Tafjord, Lambert, Gu, Arora, Bhagia, Schwenk,
  Wadden, Wettig, Hui, Dettmers, Kiela, Farhadi, Smith, Koh, Singh, and
  Hajishirzi]{muennighoff2024olmoe}
Niklas Muennighoff, Luca Soldaini, Dirk Groeneveld, Kyle Lo, Jacob Morrison,
  Sewon Min, Weijia Shi, Evan~Pete Walsh, Oyvind Tafjord, Nathan Lambert,
  Yuling Gu, Shane Arora, Akshita Bhagia, Dustin Schwenk, David Wadden,
  Alexander Wettig, Binyuan Hui, Tim Dettmers, Douwe Kiela, Ali Farhadi,
  Noah~A. Smith, Pang~Wei Koh, Amanpreet Singh, and Hannaneh Hajishirzi.
\newblock {OLMoE}: Open mixture-of-experts language models.
\newblock In \emph{The Thirteenth International Conference on Learning
  Representations}, 2025.
\newblock URL \url{https://openreview.net/forum?id=xXTkbTBmqq}.

\bibitem[Nie et~al.(2025)Nie, Zhu, You, Zhang, Ou, Hu, Zhou, Lin, Wen, and
  Li]{nie2025llada}
Shen Nie, Fengqi Zhu, Zebin You, Xiaolu Zhang, Jingyang Ou, Jun Hu, Jun Zhou,
  Yankai Lin, Ji-Rong Wen, and Chongxuan Li.
\newblock Large language diffusion models.
\newblock In \emph{Advances in Neural Information Processing Systems
  (NeurIPS)}, 2025.

\bibitem[Peng et~al.(2026)Peng, Fox, Zhang, and Tong]{peng2026repralign}
Fred~Zhangzhi Peng, Alexis Fox, Anru~R. Zhang, and Alexander Tong.
\newblock Don't retrain, align: Adapting autoregressive {LMs} to diffusion
  {LMs} via representation alignment, 2026.
\newblock URL \url{https://arxiv.org/abs/2605.06885}.

\bibitem[{Radical Numerics Inc.}(2025)]{rnd1report2025}
{Radical Numerics Inc.}
\newblock Training diffusion language models at scale using autoregressive
  models.
\newblock Technical report, Radical Numerics, October 2025.
\newblock URL \url{https://www.radicalnumerics.ai/assets/rnd1_report.pdf}.
\newblock RND1-Base-0910; code at github.com/RadicalNumerics/RND1.

\bibitem[Rajpurkar et~al.(2016)Rajpurkar, Zhang, Lopyrev, and
  Liang]{rajpurkar2016squad}
Pranav Rajpurkar, Jian Zhang, Konstantin Lopyrev, and Percy Liang.
\newblock {SQuAD}: 100,000+ questions for machine comprehension of text.
\newblock In \emph{Proceedings of EMNLP 2016}, 2016.

\bibitem[Reda et~al.(2026)Reda, Kamalu, Waleffe, Patwary, Shoeybi, and
  Catanzaro]{reda2026twotower}
Fitsum Reda, John Kamalu, Roger Waleffe, Mostofa Patwary, Mohammad Shoeybi, and
  Bryan Catanzaro.
\newblock {Nemotron-Labs-TwoTower}: Diffusion language modeling with pretrained
  autoregressive context, 2026.
\newblock URL \url{https://arxiv.org/abs/2606.26493}.

\bibitem[Sahoo et~al.(2025)Sahoo, Deschenaux, Gokaslan, Wang, Chiu, and
  Kuleshov]{sahoo2025duo}
Subham Sahoo, Justin Deschenaux, Aaron Gokaslan, Guanghan Wang, Justin Chiu,
  and Volodymyr Kuleshov.
\newblock The diffusion duality.
\newblock In \emph{ICML}, 2025.
\newblock arXiv:2506.10892.

\bibitem[Sahoo et~al.(2024)Sahoo, Arriola, Gokaslan, Marroquin, Rush, Schiff,
  Chiu, and Kuleshov]{sahoo2024mdlm}
Subham~Sekhar Sahoo, Marianne Arriola, Aaron Gokaslan, Edgar~Mariano Marroquin,
  Alexander~M Rush, Yair Schiff, Justin~T Chiu, and Volodymyr Kuleshov.
\newblock Simple and effective masked diffusion language models.
\newblock In \emph{The Thirty-eighth Annual Conference on Neural Information
  Processing Systems}, 2024.
\newblock URL \url{https://openreview.net/forum?id=L4uaAR4ArM}.

\bibitem[Shaul et~al.(2025)Shaul, Gat, Havasi, Severo, Sriram, Holderrieth,
  Karrer, Lipman, and Chen]{shaul2024kinetic}
Neta Shaul, Itai Gat, Marton Havasi, Daniel Severo, Anuroop Sriram, Peter
  Holderrieth, Brian Karrer, Yaron Lipman, and Ricky T.~Q. Chen.
\newblock Flow matching with general discrete paths: A kinetic-optimal
  perspective.
\newblock In \emph{ICLR}, 2025.
\newblock arXiv:2412.03487.

\bibitem[Tjong Kim~Sang \& De~Meulder(2003)Tjong Kim~Sang and
  De~Meulder]{tjong2003conll}
Erik~F. Tjong Kim~Sang and Fien De~Meulder.
\newblock Introduction to the {CoNLL}-2003 shared task: Language-independent
  named entity recognition.
\newblock In \emph{Proceedings of CoNLL-2003}, 2003.

\bibitem[von R{\"u}tte et~al.(2025)von R{\"u}tte, Fluri, Ding, Orvieto,
  Sch{\"o}lkopf, and Hofmann]{vonrutte2025gidd}
Dimitri von R{\"u}tte, Janis Fluri, Yuhui Ding, Antonio Orvieto, Bernhard
  Sch{\"o}lkopf, and Thomas Hofmann.
\newblock Generalized interpolating discrete diffusion.
\newblock In \emph{Proceedings of the 42nd International Conference on Machine
  Learning (ICML)}, volume 267 of \emph{Proceedings of Machine Learning
  Research}, pp.\  61810--61843, 2025.

\bibitem[Wang et~al.(2024)Wang, Ma, Zhang, Ni, Chandra, Guo, Ren, Arulraj, He,
  Jiang, Li, Ku, Wang, Zhuang, Fan, Yue, and Chen]{wang2024mmlupro}
Yubo Wang, Xueguang Ma, Ge~Zhang, Yuansheng Ni, Abhranil Chandra, Shiguang Guo,
  Weiming Ren, Aaran Arulraj, Xuan He, Ziyan Jiang, Tianle Li, Max Ku, Kai
  Wang, Alex Zhuang, Rongqi Fan, Xiang Yue, and Wenhu Chen.
\newblock {MMLU-Pro}: A more robust and challenging multi-task language
  understanding benchmark.
\newblock In \emph{Advances in Neural Information Processing Systems 37
  (NeurIPS 2024), Datasets and Benchmarks Track}, 2024.

\bibitem[Wu et~al.(2025)Wu, Zhang, Xue, Diao, Fu, Liu, Molchanov, Luo, Han, and
  Xie]{wu2025fastdllm2}
Chengyue Wu, Hao Zhang, Shuchen Xue, Shizhe Diao, Yonggan Fu, Zhijian Liu,
  Pavlo Molchanov, Ping Luo, Song Han, and Enze Xie.
\newblock {Fast-dLLM} v2: Efficient block-diffusion {LLM}, 2025.
\newblock URL \url{https://arxiv.org/abs/2509.26328}.

\bibitem[Wu et~al.(2026)Wu, Zheng, Xie, Ye, Gao, Gong, Feng, Li, Bi, Zhou, and
  Kong]{wu2026dreamon}
Zirui Wu, Lin Zheng, Zhihui Xie, Jiacheng Ye, Jiahui Gao, Shansan Gong, Yansong
  Feng, Zhenguo Li, Wei Bi, Guorui Zhou, and Lingpeng Kong.
\newblock {DreamOn}: Diffusion language models for code infilling beyond
  fixed-size canvas.
\newblock \emph{arXiv preprint arXiv:2602.01326}, 2026.

\bibitem[Yang et~al.(2025)Yang, Li, Yang, Zhang, Hui, Zheng, Yu, Gao, Huang,
  Lv, Zheng, Liu, Zhou, Huang, Hu, Ge, Wei, Lin, Tang, Yang, Tu, Zhang, Yang,
  Yang, Zhou, Zhou, Lin, Dang, Bao, Yang, Yu, Deng, Li, Xue, Li, Zhang, Wang,
  Zhu, Men, Gao, Liu, Luo, Li, Tang, Yin, Ren, Wang, Zhang, Ren, Fan, Su,
  Zhang, Zhang, Wan, Liu, Wang, Cui, Zhang, Zhou, and Qiu]{yang2025qwen3}
An~Yang, Anfeng Li, Baosong Yang, Beichen Zhang, Binyuan Hui, Bo~Zheng, Bowen
  Yu, Chang Gao, Chengen Huang, Chenxu Lv, Chujie Zheng, Dayiheng Liu, Fan
  Zhou, Fei Huang, Feng Hu, Hao Ge, Haoran Wei, Huan Lin, Jialong Tang, Jian
  Yang, Jianhong Tu, Jianwei Zhang, Jianxin Yang, Jiaxi Yang, Jing Zhou,
  Jingren Zhou, Junyang Lin, Kai Dang, Keqin Bao, Kexin Yang, Le~Yu, Lianghao
  Deng, Mei Li, Mingfeng Xue, Mingze Li, Pei Zhang, Peng Wang, Qin Zhu, Rui
  Men, Ruize Gao, Shixuan Liu, Shuang Luo, Tianhao Li, Tianyi Tang, Wenbiao
  Yin, Xingzhang Ren, Xinyu Wang, Xinyu Zhang, Xuancheng Ren, Yang Fan, Yang
  Su, Yichang Zhang, Yinger Zhang, Yu~Wan, Yuqiong Liu, Zekun Wang, Zeyu Cui,
  Zhenru Zhang, Zhipeng Zhou, and Zihan Qiu.
\newblock Qwen3 technical report, 2025.
\newblock URL \url{https://arxiv.org/abs/2505.09388}.

\bibitem[Ye et~al.(2025)Ye, Xie, Zheng, Gao, Wu, Jiang, Li, and
  Kong]{ye2025dream}
Jiacheng Ye, Zhihui Xie, Lin Zheng, Jiahui Gao, Zirui Wu, Xin Jiang, Zhenguo
  Li, and Lingpeng Kong.
\newblock {Dream 7B}: Diffusion large language models.
\newblock \emph{arXiv preprint arXiv:2508.15487}, 2025.

\bibitem[Zhu et~al.(2025)Zhu, You, Xing, Huang, Liu, Zhuang, Lu, Wang, Wang,
  Wei, Guo, Hu, Ye, Chen, Li, Tang, Feng, Hu, Zhou, Zhang, Lan, Zhao, Zheng,
  Li, Li, and Wen]{zhu2025lladamoe}
Fengqi Zhu, Zebin You, Yipeng Xing, Zenan Huang, Lin Liu, Yihong Zhuang,
  Guoshan Lu, Kangyu Wang, Xudong Wang, Lanning Wei, Hongrui Guo, Jiaqi Hu,
  Wentao Ye, Tieyuan Chen, Chenchen Li, Chengfu Tang, Haibo Feng, Jun Hu, Jun
  Zhou, Xiaolu Zhang, Zhenzhong Lan, Junbo Zhao, Da~Zheng, Chongxuan Li,
  Jianguo Li, and Ji-Rong Wen.
\newblock {LLaDA-MoE}: A sparse {MoE} diffusion language model.
\newblock \emph{arXiv preprint arXiv:2509.24389}, 2025.

\end{thebibliography}
